\documentclass[11pt]{article}

\usepackage{fullpage}
\usepackage{algorithmic}
\usepackage{algorithm}
\usepackage{graphicx,subcaption} 
\usepackage{amsfonts,amssymb,amsmath,amsthm,amsopn}	
\usepackage{hhline,booktabs,colortbl,multirow,tabularx,threeparttable} 

\usepackage{enumerate}
\usepackage{footnote}
\usepackage{float}
\usepackage{hyperref}
\usepackage{prettyref}
\usepackage{cite}
\usepackage{setspace}
\usepackage{color}
\usepackage{xcolor}  
\usepackage{titlesec} 
\usepackage{reporttitleblock} 
\usepackage{geometry}

\usepackage{tikz}
\usepackage{pgfplots}
\usetikzlibrary{positioning,shapes,shadows,arrows,calc}
\tikzstyle{component}=[rectangle, draw=black, rounded corners, fill=blue!40, drop shadow, text centered, anchor=north, text=white, minimum height=1cm]
\tikzstyle{arrow}=[->, thick]

\pgfplotsset{compat=1.18}
\usetikzlibrary{intersections}
\usetikzlibrary{pgfplots.statistics}
\usepgfplotslibrary{fillbetween}

\usepackage{eso-pic}   
\usepackage{titling}   
\usepackage{fancyhdr}  
\usepackage{lastpage}  
\usepackage{fontawesome5} 

\newcommand{\PageCounterText}{%
  {\small\color{headerblue}\faBookOpen~~\textbf{\thepage}/\pageref{LastPage}}%
}
\newcommand{\FirstPageFooterLeft}{%
  \small{\color{headerblue}\faIcon[regular]{book}}~~%
  {\small\color{headerblue}\textbf{\ReportTitleMarkText\ | \today}}%
}
\fancypagestyle{allpagesfooter}{%
  \fancyhf{}%
  \fancyfoot[R]{\PageCounterText}%
}
\fancypagestyle{firstpagefooter}{%
  \fancyhf{}%
  \fancyfoot[L]{\FirstPageFooterLeft}%
  \fancyfoot[R]{\PageCounterText}%
}
\definecolor{myblue}{RGB}{34,31,217}
\definecolor{mycyan}{gray}{.7}
\definecolor{Gray}{gray}{0.9}
\definecolor{headerblue}{RGB}{0,85,150}    
\definecolor{natureblue}{RGB}{11,61,145}   
\definecolor{titlegray}{RGB}{80,80,80}     
\definecolor{titlebox}{RGB}{245,248,252}   
\newlength{\abstractpad}
\hypersetup{
  colorlinks=true,
  linkcolor=natureblue,
  citecolor=natureblue,
  urlcolor=natureblue,
  filecolor=natureblue
}

\ReportClearAffils{}
\ReportClearAuthors{}

\ReportTitle{RIDER: 3D RNA Inverse Design with Reinforcement Learning-Guided Diffusion}
\ReportTitleMark{COLAlab Report}
\ReportAcceptedNote{Accepted as a conference paper at ICLR 2026}

\ReportAffil{1}{University of Exeter}
\ReportAffil{2}{Central South University}

\ReportAuthor{Tianmeng Hu}{1}{tianmeng0824@gmail.com}
\ReportAuthor{Yongzheng Cui}{2}{yongzhengcui@csu.edu.cn}
\ReportAuthor{Biao Luo}{2}{biao.luo@hotmail.com}
\ReportAuthor{Ke Li}{1}[corresponding]{k.li@exeter.ac.uk}

\titleformat{\section}{\color{natureblue}\normalfont\Large\bfseries}{\thesection}{1em}{}
\titleformat{\subsection}{\color{natureblue}\normalfont\large\bfseries}{\thesubsection}{1em}{}
\titleformat{\subsubsection}{\color{natureblue}\normalfont\normalsize\bfseries}{\thesubsubsection}{1em}{}

\newrefformat{fig}{Fig.~\ref{#1}}
\newrefformat{tab}{Table~\ref{#1}}
\newrefformat{sec}{Section~\ref{#1}}
\newrefformat{app}{Appendix~\ref{#1}}
\newrefformat{alg}{Algorithm~\ref{#1}}
\newrefformat{property}{Property~\ref{#1}}
\newrefformat{theorem}{Theorem~\ref{#1}}
\newrefformat{corollary}{Corollary~\ref{#1}}
\newrefformat{proposition}{Proposition~\ref{#1}}
\newrefformat{def}{Definition~\ref{#1}}
\newrefformat{eq}{equation~(\ref{#1})}

\usepackage{lscape}

\usepackage[numbers]{natbib}
\def\our{\texttt{RIDER}}

\begin{document}
\thispagestyle{firstpagefooter}

\AddToShipoutPictureBG*{%
  \AtPageUpperLeft{%
    \raisebox{-2.6cm}{\hspace{2cm}%
      \parbox[t]{\textwidth}{%
        \includegraphics[height=1.4cm]{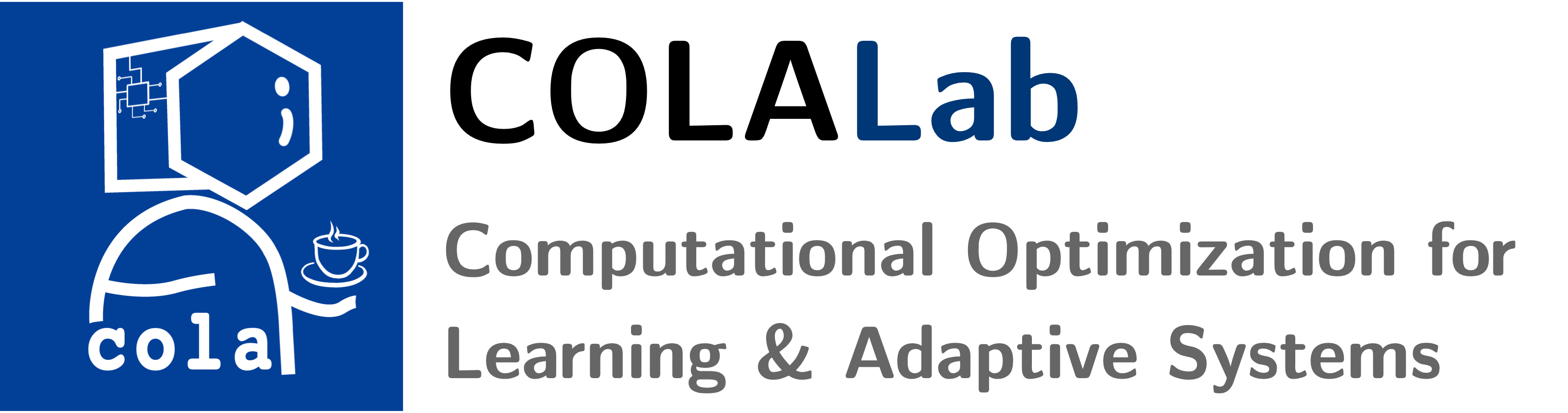}\\[-2pt]
      }%
    }%
  }%
}

\vspace*{0.8cm}

\MakeReportTitleBlock{}

\vspace{1.0em}

\noindent
	\begin{tikzpicture}
	  \node[
	    fill=titlebox,
	    rounded corners=10pt,
	    inner xsep=\abstractpad,
	    inner ysep=10pt
	  ]{%
	    \begin{minipage}{\dimexpr\textwidth-2\abstractpad\relax}
	      \setlength{\parindent}{0pt}%
%

{\centering\large\color{natureblue}\textbf{Abstract}\par}
\vspace{0.4em}
\noindent The inverse design of RNA three-dimensional (3D) structures is crucial for engineering functional RNAs in synthetic biology and therapeutics. While recent deep learning approaches have advanced this field, they are typically optimized and evaluated using native sequence recovery, which is a limited surrogate for structural fidelity, since different sequences can fold into similar 3D structures and high recovery does not necessarily indicate correct folding. To address this limitation, we propose \our, an RNA Inverse DEsign framework with Reinforcement learning that directly optimizes for 3D structural similarity. First, we develop and pre-train a GNN-based generative diffusion model conditioned on the target 3D structure, achieving a $9\%$ improvement in native sequence recovery over state-of-the-art methods. Then, we fine-tune the model with an improved policy gradient algorithm using four task-specific reward functions based on 3D self-consistency metrics. Experimental results show that \our\ improves structural similarity by over $100\%$ across all metrics and discovers designs that are distinct from native sequences.

\vspace{0.6em}
{\normalsize\textbf{\color{natureblue}Project Repo: }}
\texttt{\href{https://github.com/COLA-Laboratory/RIDER}
{github.com/COLA-Laboratory/RIDER}}

	    \end{minipage}
	  };
	\end{tikzpicture}


%

\section{Introduction}\label{sec:introduction}

Ribonucleic acid (RNA) is a fundamental biomolecule with diverse biological functions, such as catalysis, gene regulation, and metabolite sensing~\citep{serganov2007ribozymes}. RNA function is closely linked to its complex three-dimensional (3D) structure~\citep{zhang2022advances}. The RNA inverse design problem~\citep{hofacker1994fast} – finding a nucleotide sequence that will fold into a desired target structure – is therefore critical for designing RNAs with tailored functions for therapeutics and synthetic biology~\citep{TaftPMDM10,SerganovN13,DykstraKS22,Guo10}.

RNA structure is hierarchical, progressing from the primary nucleotide sequence to secondary structures formed by local base pairing and culminating in the tertiary 3D structure, which involves long-range interactions, both canonical Watson-Crick and non-canonical pairs. While most computational RNA inverse design target RNA secondary structure~\citep{AndronescuFHHC04, BuschB06, MartinCD13, MartinCD15, kleinkauf2015antarna,eastman2018solving,runge2019learning,zhou2023rna}, achieving precise biological function often requires control over the final 3D structure, as secondary structure only partially dictates the tertiary fold~\citep{huang2024ribodiffusion}. Consequently, recent research has focused on tertiary structure-based inverse design. State-of-the-art methods employ deep learning, using graph neural networks (GNNs)~\citep{joshi2025grnade,wong2024deep,tan2024rdesign} or generative diffusion models~\citep{huang2024ribodiffusion}. These approaches are inspired by successes in protein design~\citep{dauparas2022robust}, demonstrate significant improvements in speed and performance over traditional physically-based method~\citep{leman2020macromolecular}. However, they are typically optimized and evaluated based on their ability to recover the native sequence for a given structure, measured by native sequence recovery (NSR)~\citep{joshi2025grnade}. 

\begin{figure*}[t!]
\center
    \includegraphics[width=0.99\linewidth]{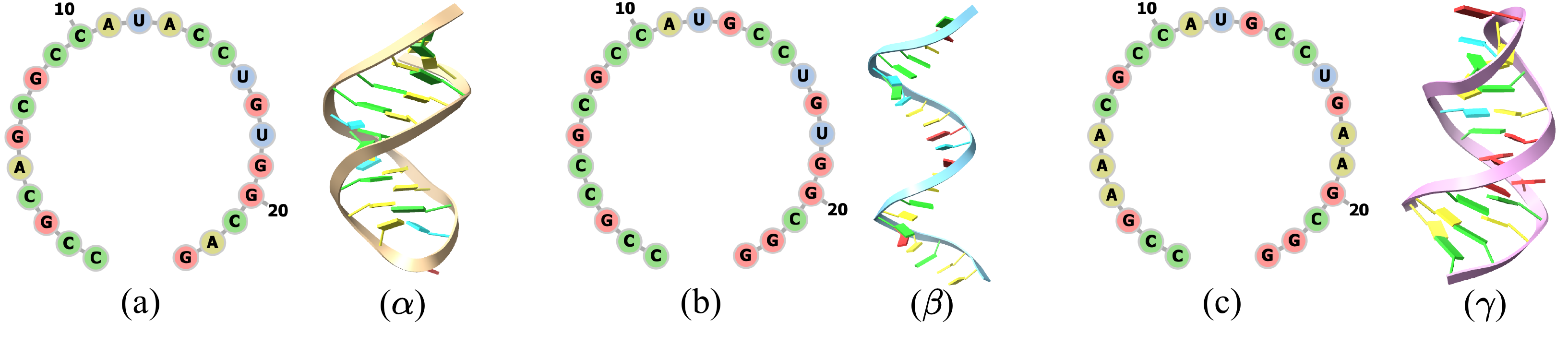}
    \caption{Visualization of sequences (a), (b), and (c), and their corresponding 3D structures ($\alpha$), ($\beta$), and ($\gamma$) predicted by RhoFold~\citep{shen2024accurate}. Although sequences (a) and (b) differ by only 3 nucleotides, and (b) and (c) by 5 nucleotides, their folded structures exhibit clear differences.}
    \label{fig:example}
\end{figure*}

Optimizing solely for NSR is problematic for RNA design. Unlike proteins, the relationship between RNA sequence and structure is highly degenerate: multiple distinct sequences can fold into similar structures~\citep{assmann2023rock}. Furthermore, similar sequences do not necessarily result in similar structures, as illustrated in Figure~\ref{fig:example}. The true objective of RNA inverse design is structural fidelity – finding any sequence that folds correctly – not necessarily recovering the single native sequence observed in the training data. Relying on NSR as a proxy objective has limitations: 1) It does not directly optimize for structural similarity; improving sequence similarity to the native sequence does not ensure the designed structure matches the target~\citep{liu2025sentences}. Notably, most secondary structure design methods avoid NSR, instead optimizing structural distance~\citep{kleinkauf2015antarna,eastman2018solving,runge2019learning,zhou2023rna}. 2) A key problem in RNA biology is to identify RNA sequences with structural similarity to natural sequences~\citep{morandi2022shape}. Over-optimizing for NSR restricts exploration to sequences near the native one, potentially preventing the discovery of novel, non-native sequences. 

To overcome these limitations, we propose \our, a novel approach that directly optimizes structural similarity using reinforcement learning (RL), integrating the strengths of generative models with the targeted optimization capabilities of RL. We first pre-train a generative diffusion model on RNA structure datasets, optimizing for NSR to enable the model to learn 3D structural representations and capture sequence-structure relationships. Subsequently, we fine-tune this model using RL, employing a reward function based on the structural similarity between the predicted structure folded from the generated sequence and the target structure. This RL phase allows exploration of the vast sequence space, potentially uncovering novel sequences far from the native one in the dataset. Conceptually, this aligns with recent successes in using RL to fine-tune large generative models for specific objectives~\citep{black2024training}.

Our main contributions are:
\begin{itemize}

    \item We propose the first reinforcement learning framework for RNA 3D inverse design, which directly optimizes structural similarity rather than relying on surrogate sequence recovery objectives.
    
    \item We develop and pre-train \texttt{RIDE}, a generative diffusion model for RNA inverse design conditioned on the target structural information, achieving an NSR of $61\%$ and surpassing the strongest baseline at $56\%$.

    \item We improve existing RL algorithms for fine-tuning diffusion models and apply them to the pre-trained \texttt{RIDE}, resulting in the complete \our\ framework. Specifically, we adopt a batch-mean baseline for advantage estimation and further incorporate a moving average strategy to stabilize training. We design four novel reward functions based on 3D structural self-consistency metrics to guide RL fine-tuning. 
    
    \item Through experiments, we demonstrate that \our\ achieves over a $100\%$ improvement over existing state-of-the-art methods across all three 3D self-consistency metrics. 
\end{itemize}



\section{Related work}
\label{sec:Related Work}

\paragraph{RNA inverse design.}
Computational RNA inverse design seeks sequences that fold into a desired target structure. Most existing methods primarily focus on secondary structure. These approaches employ RNA secondary folding prediction tools~\citep{lorenz2011viennarna,MarkhamZ08,AliMM23} to compute the structural distance between the designed and target structures, and then search for sequences that meet the design objectives. Methods in this category include local search~\citep{HofackerFSBTS94, AndronescuFHHC04, BuschB06}, metaheuristic algorithms~\citep{Taneda12, TaheriGM14, TaheriG15, kleinkauf2015antarna, zhou2023rna}, and reinforcement learning~\citep{eastman2018solving, runge2019learning}. Recent efforts have shifted toward 3D structure design. Physics-based approaches such as Rosetta~\citep{leman2020macromolecular} support fixed-backbone design, but are computationally expensive. More recent methods incorporate deep learning: RiboDiffusion~\citep{huang2024ribodiffusion} employs generative diffusion models with Transformer components. RDesign~\citep{tan2024rdesign}, RhoDesign~\citep{wong2024deep}, R3Design~\citep{tan2025r3design}, and gRNAde~\citep{joshi2025grnade} leverage GNNs to extract 3D structural representations and train generative models to recover sequences from structural inputs, achieving high sequence recovery.
A common thread among these state-of-the-art methods is their reliance on supervised learning maximizing sequence recovery, which we argue is an indirect and potentially suboptimal proxy for structural fidelity. Our work addresses this by introducing an RL framework for direct structural optimization.

\paragraph{RNA structure prediction.} 
Since experimentally determining RNA structures is labor-intensive and expensive, computational prediction methods have become indispensable~\citep{piao2017rna}. Early approaches such as RNAComposer~\citep{popenda2012automated} assembled structures from libraries of known fragments. More recently, deep learning has become dominant, leveraging statistical patterns learned from existing RNA structures. Representative examples include trRosettaRNA~\citep{wang2023trrosettarna}, DRFold~\citep{li2023integrating}, and RhoFold~\citep{shen2024accurate}. Although RNA structure prediction is essential, progress in RNA functionality and drug discovery ultimately depends on methods for designing novel sequences. Nonetheless, continued improvements in prediction methods are expected to further enhance the effectiveness of our design framework.

\paragraph{Reinforcement learning.} 
Reinforcement Learning provides a framework for optimizing policies toward specific goals defined by reward functions and has been widely applied to solve dynamic decision-making problems~\citep{mnih2015human, lillicrap2015continuous, rashid2020monotonic, hu2023mo}. RL has achieved remarkable success across a range of challenging domains, including competitive games and multi-agent environments~\citep{lowe2017multi, vinyals2019grandmaster}, Go~\citep{silver2016mastering, silver2017mastering}, robotic control~\citep{haarnoja2024learning, hu2024pa2d}, and autonomous drone racing~\citep{kaufmann2023champion}, with several systems attaining or surpassing human-expert performance. Recently, RL has been employed to fine-tune large language models to align their outputs with human intent~\citep{ouyang2022training, bai2022constitutional}, or to enhance reasoning abilities for solving complex mathematical and programming tasks~\citep{guo2025deepseek, wang2024enhancing}. In addition, recent works apply RL to fine-tune pretrained generative models, aligning them with human preferences or optimizing them for specific objectives~\citep{fan2023dpok, zhang2024large, black2024training}. For example, DDPO~\citep{black2024training} designs reward functions for text-to-image generation, enabling diffusion models to adapt toward targets such as image compressibility or aesthetic quality.


\section{Method}
\label{sec:method}
Our approach for 3D RNA inverse design is formulated as a conditional generative process based on a diffusion model. Given a target RNA 3D backbone structure, the model learns to generate a compatible RNA sequence. The method comprises two main stages: first, a geometric graph representation of the RNA backbone is extracted and processed by a structure encoder to obtain a conditional embedding; second, a diffusion model, conditioned on this structural embedding, iteratively refines a noisy sequence representation to produce the final RNA sequence. Building upon this, we further introduce a reinforcement learning fine-tuning stage to directly optimize the structural similarity of the generated sequences. The pseudocode of the algorithms is provided in Appendix~\ref{sec:appendix_algorithm}.

\subsection{Tertiary structure representation}
\label{subsec:structure_representation}

To capture the geometric properties of RNA tertiary structures, we represent each molecule as a geometric graph, where nodes correspond to nucleotides and edges encode spatial proximity. A structural encoder based on GVP-GNN~\citep{jing2021learning} processes this graph to produce node-level embeddings that are equivariant to 3D rotations and translations. These embeddings compactly summarize the local and global geometry of the RNA backbone and serve as conditioning context for the diffusion model. Details of the structural encoder are provided in Appendix~\ref{sec:appendix_representation}. Figure~\ref{fig:pipeline} illustrates the overall pipeline.

\begin{figure}[t]
    \centering
    \includegraphics[width=1.0\linewidth]{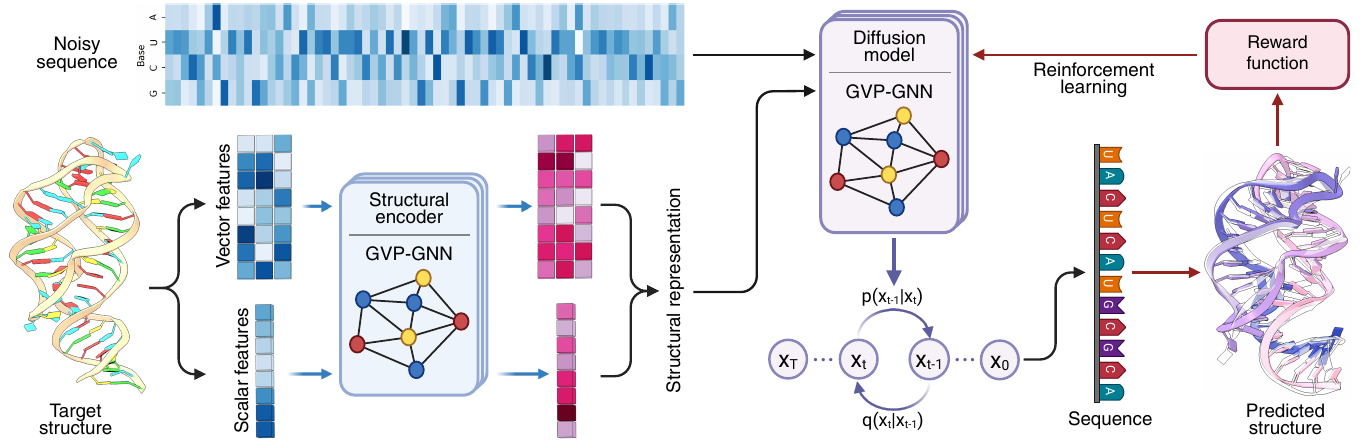}
    \caption{Overview of the \our\ framework. RNA tertiary structures are processed by a GVP-GNN encoder to produce structural embeddings. These embeddings condition the diffusion model for sequence generation, which is further optimized by RL to maximize structural similarity.}
    \label{fig:pipeline}
\end{figure}

\subsection{Conditional diffusion model for RNA sequence generation}
\label{subsec:diffusion_pretrain}

We formulate RNA sequence design as learning a conditional distribution $p(\mathbf{x}_0 \mid \mathbf{h}_c)$, where $\mathbf{x}_0$ denotes the one-hot encoded sequence ($N$ nucleotides, $C=4$ bases, $\mathbf{x}_0 \in \{0,1\}^{N \times C}$) and $\mathbf{h}_c$ is the structural embedding from the encoder. To model this distribution, we adopt a variance-preserving diffusion probabilistic model \citep{ho2020denoising, song2021scorebased}, which learns to predict the noise added to a clean sequence. The forward diffusion process gradually adds Gaussian noise to the clean data $\mathbf{x}_0$ over a continuous time variable $t \in [\epsilon_{\texttt{time}}, T]$, where $\epsilon_{\texttt{time}}$ is a small positive constant and $T$ is the final time. The noised sample $\mathbf{x}_t$ at time $t$ is given by:
\begin{equation}
    \mathbf{x}_t = \alpha_t \mathbf{x}_0 + \sigma_t \varepsilon,
    \label{eq:forward_process}
\end{equation}
where $\varepsilon \sim \mathcal{N}(0, I)$, and $\alpha_t, \sigma_t$ are schedule functions derived from a predefined noise schedule $\beta_t$. Specifically, $\alpha_t^2 = \exp\left(-\int_0^t \beta_s ds\right)$ and $\sigma_t^2 = 1 - \alpha_t^2$. The noise level for a given $t$ is often represented as $\lambda_t = \log(\alpha_t^2 / \sigma_t^2)$.

The core of our generative model is a noise prediction network $\epsilon_{\theta}(\mathbf{x}_t, t, \mathbf{h}_c)$ with parameters $\theta$. This network is trained to predict the noise $\varepsilon$ from the noisy input $\mathbf{x}_t$, conditioned on the time $t$ and the structural context $\mathbf{h}_c$. The network $\epsilon_{\theta}$ is composed of $L_D=5$ GVP-GNN layers.
The input to this noise prediction network for each node $i$ combines its noisy sequence representation $(\mathbf{x}_t)_i$, an embedding of the time step $t_{\texttt{emb}}$, and its structural embedding $(\mathbf{h}_c)_i$. This combined representation is then processed by the network to predict the noise component $(\hat{\varepsilon})_i$ for each nucleotide.
The model is trained by minimizing the mean squared error between the true noise $\varepsilon$ and the predicted noise $\hat{\varepsilon}_{\theta}$:
\begin{equation}
    \mathcal{L}_{\texttt{pretrain}}(\theta) = \mathbb{E}_{t, \mathbf{x}_0, \varepsilon, \mathbf{h}_c} \left[ || \varepsilon - \epsilon_{\theta}(\alpha_t \mathbf{x}_0 + \sigma_t \varepsilon, t, \mathbf{h}_c) ||^2 \right].
    \label{eq:loss_pretrain}
\end{equation}
During inference, we generate sequences by reversing the diffusion process. Starting from a random Gaussian noise sample $\mathbf{x}_T \sim \mathcal{N}(0, I)$, we iteratively denoise it using the Denoising Diffusion Implicit Models (DDIM) \citep{song2021denoising} sampler. After $N_{\texttt{steps}}$ iterations, we obtain an estimate of the clean sequence $\hat{\mathbf{x}}_0$. The final discrete RNA sequence is obtained by applying an argmax operation over the $C$ channels for each nucleotide in $\hat{\mathbf{x}}_0$.

\subsection{Reinforcement learning for structural similarity optimization}
\label{subsec:rl_finetune}

Although the pre-trained diffusion model can learn sequence-structure relationships, its training objective does not directly maximize the structural similarity between the folded structure of the generated sequence and the target structure. To overcome this limitation, we employ a reinforcement learning-based method to fine-tune the pre-trained diffusion model, inspired by the denoising diffusion policy optimization (DDPO) framework \citep{black2024training}, to directly optimize for rewards related to downstream tasks.

We frame the denoising sampling process of the diffusion model as a Markov decision process (MDP). In this MDP:
\begin{itemize}
    \item \textbf{State $s_t$}: Defined as $(\mathbf{x}_t, t, \mathbf{h}_c)$, representing the current noisy sequence, time step, and structural condition.
    \item \textbf{Action $a_t$}: Defined as the transition from $\mathbf{x}_t$ to $\mathbf{x}_{t-\Delta t}$ in one denoising step. In our context, this corresponds to the noise (or equivalently, the predicted $\hat{\mathbf{x}}_0$) predicted by the model $\epsilon_{\theta}$ given $s_t$, which determines the mean of the next state $\mathbf{x}_{t-\Delta t}$.
    \item \textbf{Policy $\pi_{\theta}(a_t|s_t)$}: Parameterized by the diffusion model $\epsilon_{\theta}$.
    \item \textbf{Reward $R(\hat{\mathbf{x}}_0, \mathbf{h}_c^\texttt{target})$}: Obtained only at the end of the trajectory (i.e., after generating the complete $\hat{\mathbf{x}}_0$).
\end{itemize}

\paragraph{Advantage estimation}
We employ a policy gradient method to fine-tune the diffusion model. The core idea is to update the model parameters $\theta$ using the complete denoising trajectories $\{\mathbf{x}_T, \mathbf{x}_{T-1}, ..., \hat{\mathbf{x}}_0\}$ and their final rewards $R_{traj} = R(\hat{\mathbf{x}}_0, \mathbf{h}_c^\texttt{target})$. The original DDPO applies an importance sampling (IS) based estimator:
\begin{equation}
\nabla_\theta\mathcal{J}_{\texttt{DDRL}}=\mathbb{E}_{\tau \sim p(\tau|\pi_{\theta_\texttt{old}})}\left[\sum_{k=0}^{N_\texttt{steps}-1}\frac{\pi_\theta(a_k|s_k)}{\pi_{\theta_{\texttt{old}}}(a_k|s_k)}\nabla_\theta\log \pi_\theta(a_k|s_k)R_\texttt{traj}\right],
\label{eq:ddpo_is_reward}
\end{equation}
where $a_k$ represents the action of transitioning from state $s_k$ (i.e., $(\mathbf{x}_{t_k}, t_k, \mathbf{h}_c)$) to $\mathbf{x}_{t_{k-1}}$ at time step $k$, $\pi_\theta(a_k|s_k)$ is the probability of this action under the current policy, and $\pi_{\theta_\texttt{old}}$ is the old policy used for sampling.
While this approach is effective in image generation, we found that directly using the reward $R_\texttt{traj}$ in our RNA design context leads to high variance and instability during optimization. Therefore, we replace the reward function with an advantage function $A(\hat{\mathbf{x}}_0, \mathbf{h}_c^\texttt{target})$ to reduce the variance of gradient estimates. Specifically, for each target structure, we collect a batch of experience trajectories by running the denoising sampling process multiple times. Each trajectory includes a sequence of latent states $\mathbf{x}_t$, the log-probability $\log \pi_{\theta_\texttt{old}}(a_t|s_t)$ for each step, and the final structural similarity reward $R_\texttt{traj}$. We then compute the average reward over this batch as a baseline $b$:
\begin{equation}
    b = \mathbb{E}_{\tau \sim p(\tau|\pi_{\theta_\texttt{old}})} [R_\texttt{traj}].
    \label{eq:baseline}
\end{equation}
However, we observe in our experiments that using a baseline based solely on the batch average can lead to instability. Due to the stochasticity of the sampling process, baseline values computed from adjacent batches may vary significantly, resulting in large fluctuations in advantage estimates and hindering stable learning. Increasing the batch size may mitigate this issue but at the cost of reduced training efficiency. 
To address this, we adopt a moving average strategy for baseline estimation. Let $b^{(i)}$ denote the moving average baseline after the $i$-th batch of experience collection, and let $\bar{R}^{(i)}_{\texttt{batch}}$ be the average reward of the $i$-th batch. The baseline is updated according to:
\begin{equation}
b^{(i)} = \beta_{\texttt{baseline}} \cdot b^{(i-1)} + (1 - \beta_{\texttt{baseline}}) \cdot \bar{R}^{(i)}_{\texttt{batch}},
\end{equation}
where $b^{(0)}$ is initialized as the average reward of the first batch, and $\beta_{\text{baseline}}$ is a smoothing factor that controls the update rate. The smoothed baseline $b \equiv b^{(i)}$ is then used in advantage computation\footnote{Appendix~\ref{sec:adv_ablation} investigates the impact of advantage function estimation.}.
The advantage function is then defined as $A(\hat{\mathbf{x}}_0, \mathbf{h}_c^\texttt{target}) = R_\texttt{traj} - b$, and the policy gradient estimator based on advantage is correspondingly modified to:
\begin{equation}
\nabla_\theta\mathcal{J}_{\texttt{RL}}=\mathbb{E}_{\tau \sim p(\tau|\pi_{\theta_\texttt{old}})}\left[\left(\sum_{k=0}^{N_\texttt{steps}-1}\frac{\pi_\theta(a_k|s_k)}{\pi_{\theta_{\texttt{old}}}(a_k|s_k)}\nabla_\theta\log \pi_\theta(a_k|s_k)\right)A(\hat{\mathbf{x}}_0, \mathbf{h}_c^\texttt{target})\right].
\label{eq:advantage_pg}
\end{equation}
During the sampling process, to ensure diversity in each batch of samples and effectively explore the policy space, we perform one deterministic sampling (temperature parameter set to $0$) and multiple stochastic samplings. In the stochastic sampling phase, we uniformly sample a temperature from a discrete set (e.g., $\{0.1, 0.3, 0.5, 0.7, 0.9\}$ with a probability of 1/5 for each) to control the randomness of the generation process. Furthermore, we adopt the clipping technique \citep{schulman2017proximal} to constrain the policy updates. We apply the importance sampling ratio $r_k(\theta) = \frac{\pi_\theta(a_k|s_k)}{\pi_{\theta_{\texttt{old}}}(a_k|s_k)}$ from Equation (\ref{eq:advantage_pg}) to the advantage function and clip the result, forming the following optimization objective:
\begin{equation}
    \mathcal{L}^{RL}(\theta) = \mathbb{E}_{\tau \sim p(\tau|\pi_{\theta_\texttt{old}})} \left[ \sum_{k=0}^{N_\texttt{steps}-1} \min(r_k(\theta) A, \text{clip}(r_k(\theta), 1-\epsilon_\texttt{clip}, 1+\epsilon_\texttt{clip}) A) \right],
    \label{eq:ppo_clip_objective_sum}
\end{equation}
where $A$ is the advantage estimation of the entire trajectory, and $\epsilon_\texttt{clip}$ is the clipping parameter. 

\paragraph{Reward function}
\label{par:reward_function}
The reward function is designed to quantify the structural similarity between the predicted 3D structure—obtained by folding the generated RNA sequence—and the target 3D structure. We use the computational structure prediction model RhoFold \citep{shen2024accurate} to predict the folded structure of the generated sequence. We employ three widely used structural similarity metrics:
\begin{itemize}
    \item \textbf{GDT\_TS}~\citep{zemla1999processing} (Global Distance Test Total Score): measures the percentage of backbone atoms within preset distance thresholds ($1$\AA, $2$\AA, $4$\AA, $8$\AA) of their counterparts after optimal superposition. It emphasizes well-aligned regions and ranges from $0$ to $1$, with higher scores indicating better structural alignment.
    \item \textbf{RMSD}~\citep{kabsch1976solution} (Root Mean Square Deviation): computes the average displacement between corresponding backbone atoms after superposition. It is sensitive to outliers; lower values indicate higher similarity.
    \item \textbf{TM-score}~\citep{zhang2004scoring} (Template Modeling Score): evaluates global fold similarity using a length-normalized function. It is tolerant to local deviations and suitable for comparing structures of different lengths. Values range from $0$ to $1$, with higher scores reflecting better agreement.
\end{itemize}
A detailed explanation of these metrics is provided in Appendix~\ref{sec:3d_metrics}. Based on these metrics, we explore various reward functions. First, we define base reward functions using individual metrics to analyze the model’s ability to optimize for specific structural features:
\begin{align}
    R^{\texttt{gdt}} &= (\text{GDT\_TS} \times w_{\texttt{gdt\_scale}})^2 \label{eq:reward_gdt}, \\
    R^{\texttt{tm}} &= (\text{TM-score} \times w_{\texttt{tm\_scale}})^2 \label{eq:reward_tm}, \\
    R^{\texttt{rmsd}} &= -(\text{RMSD} \times w_{\texttt{rmsd\_scale}})^2 \label{eq:reward_rmsd},
\end{align}
where $w_{\texttt{gdt\_scale}}$, $w_{\texttt{tm\_scale}}$, and $w_{\texttt{rmsd\_scale}}$ are scaling factors. Considering that RMSD and GDT\_TS reflect structural similarity from different perspectives, we also design a combined reward function to balance the pursuit of precise atomic matching and overall fold similarity:
\begin{equation}
    R^{\texttt{gdt\_rmsd}} = -(\text{RMSD} \times w_{\texttt{rmsd\_scale}})^2 + (\text{GDT\_TS} \times w_{\texttt{gdt\_scale}})^2.
    \label{eq:reward_combined_base}
\end{equation}
It is considered that when GDT\_TS $> 0.5$ or RMSD $< 2.0 \text{\AA}$, the predicted structure has significant similarity to the target structure, meeting an acceptable design standard \citep{tan2024rdesign, joshi2025grnade}. To encourage the model to generate sequences with high structural fidelity, we design an additional reward component $R_\texttt{bonus}$ for sequences that meet these criteria, computed as:
\begin{equation}
    R_\texttt{bonus} =
    \begin{cases}
        (\text{GDT\_TS} - \tau_{\texttt{gdt}}) \times w_{\texttt{bonus\_gdt}}, & \text{if GDT\_TS} > \tau_{\texttt{gdt}} \\
        (\tau_{\texttt{rmsd}} - \text{RMSD}) \times w_{\texttt{bonus\_rmsd}}, & \text{else if RMSD} < \tau_{\texttt{rmsd}} \\
        0, & \text{otherwise}
    \end{cases}
    \label{eq:reward_bonus}
\end{equation}
where $\tau_{\texttt{gdt}} = 0.5$ and $\tau_{\texttt{rmsd}} = 2.0$ are threshold values, and $w_{\texttt{bonus\_gdt}}$ and $w_{\texttt{bonus\_rmsd}}$ are reward scaling coefficients.
The final total reward function $R_\texttt{final}$ used in our experiments is defined as the sum of a base reward and a bonus reward:
\begin{equation}
    R_\texttt{final} = R_\texttt{base} + R_\texttt{bonus},
    \label{eq:reward_total_final}
\end{equation}
where $R_\texttt{base}$ corresponds to one of the four base reward functions defined above.


\section{Experiments}
\label{sec:experiments}

\subsection{Settings}

\paragraph{Dataset}
We use the RNA tertiary structure dataset published by \citet{joshi2025grnade} for pre-training and evaluating our model. The dataset is derived from the RNASolo repository~\citep{adamczyk2022rnasolo} and contains $4,223$ unique RNA sequences and a total of $12,011$ RNA structures. It is pre-partitioned based on structural similarity into a training set, a validation set, and a test set, with the latter two each containing $100$ samples.

\paragraph{Hyperparameters}
Both the structure encoder and the noise prediction network in the diffusion model use $5$ layers of GVP-GNNs. The initial learning rate is set to $3 \times 10^{-4}$ and is decayed by a factor of $0.9$ if the validation performance does not improve for $5$ consecutive epochs. The model is trained for a total of $150$ epochs. The overall number of parameters in our model is $10.2$ million. All GVP-GNN layers apply a dropout rate of $0.5$ to mitigate overfitting. To improve robustness, Gaussian noise with a standard deviation of $0.1$ is added to the node coordinates during training. For sequence generation, we use the DDIM sampler~\citep{song2021denoising} with $N_{\texttt{steps}} = 50$ denoising steps. 
After pre-training, we fine-tune the diffusion model using a policy gradient algorithm. The learning rate is set to $5 \times 10^{-5}$ without learning rate scheduling. Training is conducted for $80$ reinforcement learning epochs, with $2$ policy updates performed in each epoch. The batch size is $60$, meaning that $60$ experience trajectories (each corresponding to a full denoising process) are sampled per epoch. Appendix~\ref{sec:appendix_hyperparameters} provides detailed information on the model architecture and training hyperparameters.

\begin{figure*}[t!]
\center
    \includegraphics[width=1.0\linewidth]{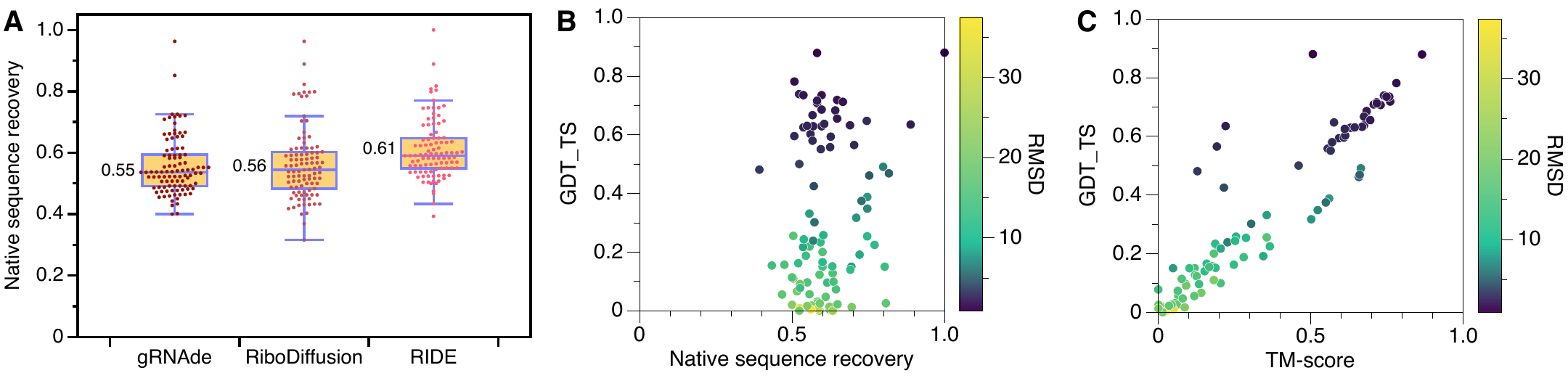}
    \caption{Results of supervised learning pre-training. \textbf{A}. Comparison of native sequence recovery on the test set. \texttt{RIDE} is compared against RiboDiffusion and gRNAde. The best NSR among $16$ designs per target (sampled at temperature $0.1$) is reported. \textbf{B}. Relationship between NSR and structural similarity (GDT\_TS and RMSD) for \texttt{RIDE} designs. Color denotes RMSD. \textbf{C}. Correlation among GDT\_TS, TM-score, and RMSD for the designed structures.}
    \label{fig:dm_results}
\end{figure*}

\subsection{Supervised learning pre-training}
We compare our proposed conditional diffusion model \texttt{RIDE} with two recent state-of-the-art (SOTA) methods: RiboDiffusion~\citep{huang2024ribodiffusion} and gRNAde~\citep{joshi2025grnade}. RiboDiffusion is a Transformer-based diffusion model, whereas gRNAde is a GNN-based generative model that employs an autoregressive decoder to generate RNA sequences.

\paragraph{\texttt{RIDE} achieves the best sequence recovery}
We evaluate the three methods on the test set by comparing their NSR. For each target structure, each method performs $16$ sampling runs at a temperature of $0.1$ to generate $16$ candidate sequences, and the best NSR among them is recorded. The results are shown in Figure~\ref{fig:dm_results}A.
\texttt{RIDE} achieves an average sequence recovery of $61\%$, representing improvements of $9\%$ and $11\%$ over RiboDiffusion and gRNAde, respectively. These results demonstrate that, even without reinforcement learning fine-tuning, our method outperforms existing SOTA baselines in terms of sequence recovery.

\paragraph{Native sequence recovery does not reflect structural similarity}
Using the same methodology as in Figure~\ref{fig:dm_results}A, we employ \texttt{RIDE} to perform inverse design on the $100$ structures in the test set. For each structure, we record the best native sequence recovery and its corresponding 3D structural similarity metrics, as shown in Figure~\ref{fig:dm_results}B.
The results reveal no clear correlation between sequence recovery and 3D structural similarity. For example, when sequence recovery is around $50\%$, the associated GDT\_TS values vary widely from $0$ to $0.9$. As sequence recovery increases, the distribution of GDT\_TS does not consistently shift toward higher values. In fact, even at high levels of sequence recovery, the resulting structures may still exhibit low similarity to the target structure.
These observations suggest that native sequence recovery alone is insufficient to assess the structural fidelity of designed sequences. This further highlights that optimizing sequence recovery is not an ideal surrogate objective for 3D RNA inverse design.

\paragraph{Relationship among three self-consistency metrics}
Figure~\ref{fig:dm_results}C illustrates the relationships among GDT\_TS, TM-score, and RMSD. GDT\_TS and TM-score show a strong correlation, with a Pearson correlation coefficient of $0.885$. A few outliers highlight differences between these metrics. For instance, some samples exhibit low RMSD values (indicated by darker colors), high GDT\_TS, but unexpectedly low TM-scores. This discrepancy arises because TM-score is more sensitive to deviations in shorter structures.
Since these three metrics capture structural similarity from different perspectives, we design distinct reward functions based on each of them during the RL fine-tuning phase to investigate their respective impact on performance.

\begin{figure*}[t!]
\center
    \includegraphics[width=1.0\linewidth]{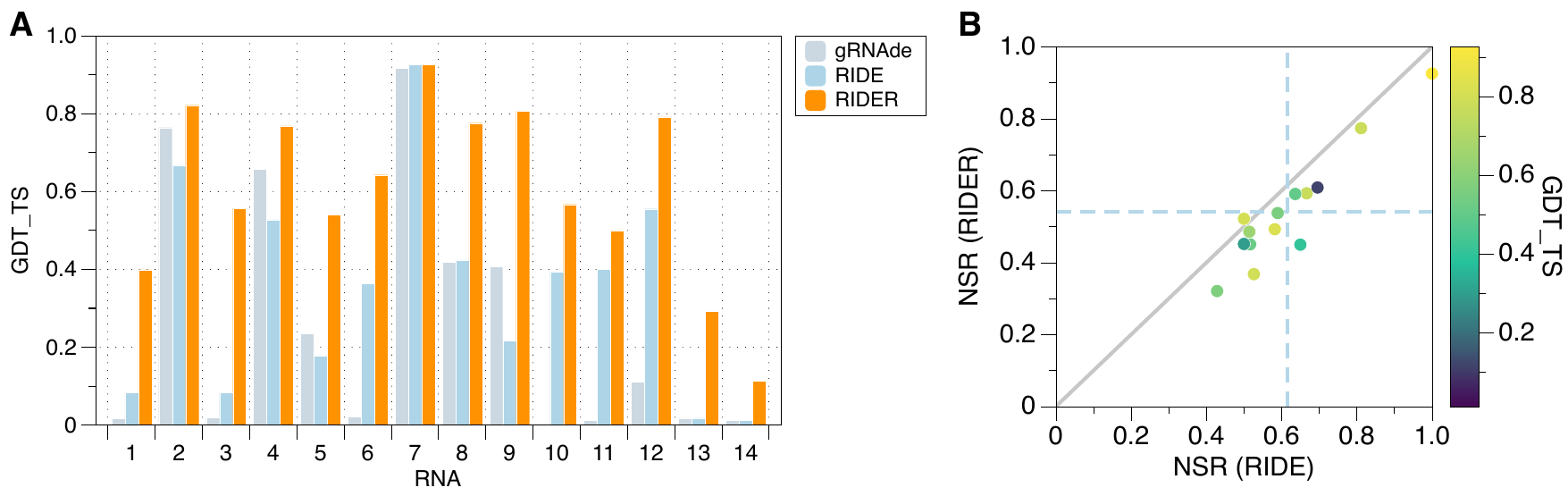}
    \caption{Results of reinforcement learning fine-tuning. \textbf{A}. GDT\_TS comparison on $14$ RNA structures of interest~\citep{das2010atomic} for gRNAde, \texttt{RIDE} (pre-trained), and \our\ (fine-tuned with $R^{\texttt{gdt\_rmsd}}$). \textbf{B}. Comparison of native sequence recovery before (\texttt{RIDE}) and after (\our) RL fine-tuning. Color indicates GDT\_TS after RL fine-tuning. The results for the other two metrics are provided in Appendix~\ref{sec:additional}.}
    \label{fig:rl_results}
\end{figure*}

\begin{figure}[tbp]
  \centering
  \begin{subfigure}[t]{0.32\textwidth}
    \centering
    \includegraphics[width=0.52\textwidth]{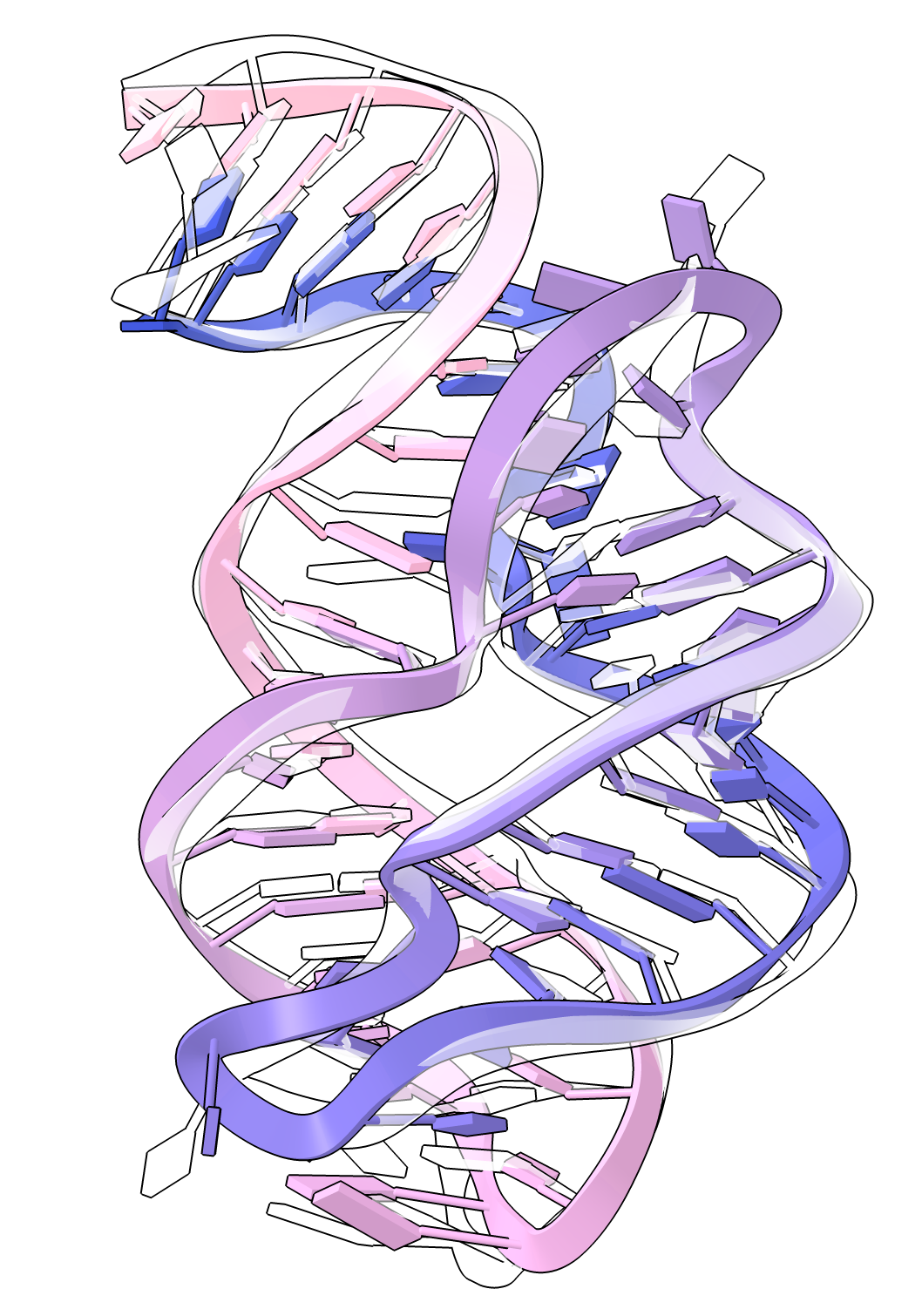}
    {\ttfamily
    \\
      GDT\_TS = 0.822 \\
      TM-score = 0.825\\
      RMSD = 1.444
    }
    \caption*{\textbf{PDB: 3GAO}}
  \end{subfigure}
  \hfill
  \begin{subfigure}[t]{0.32\textwidth}
    \centering
    \includegraphics[width=1\textwidth]{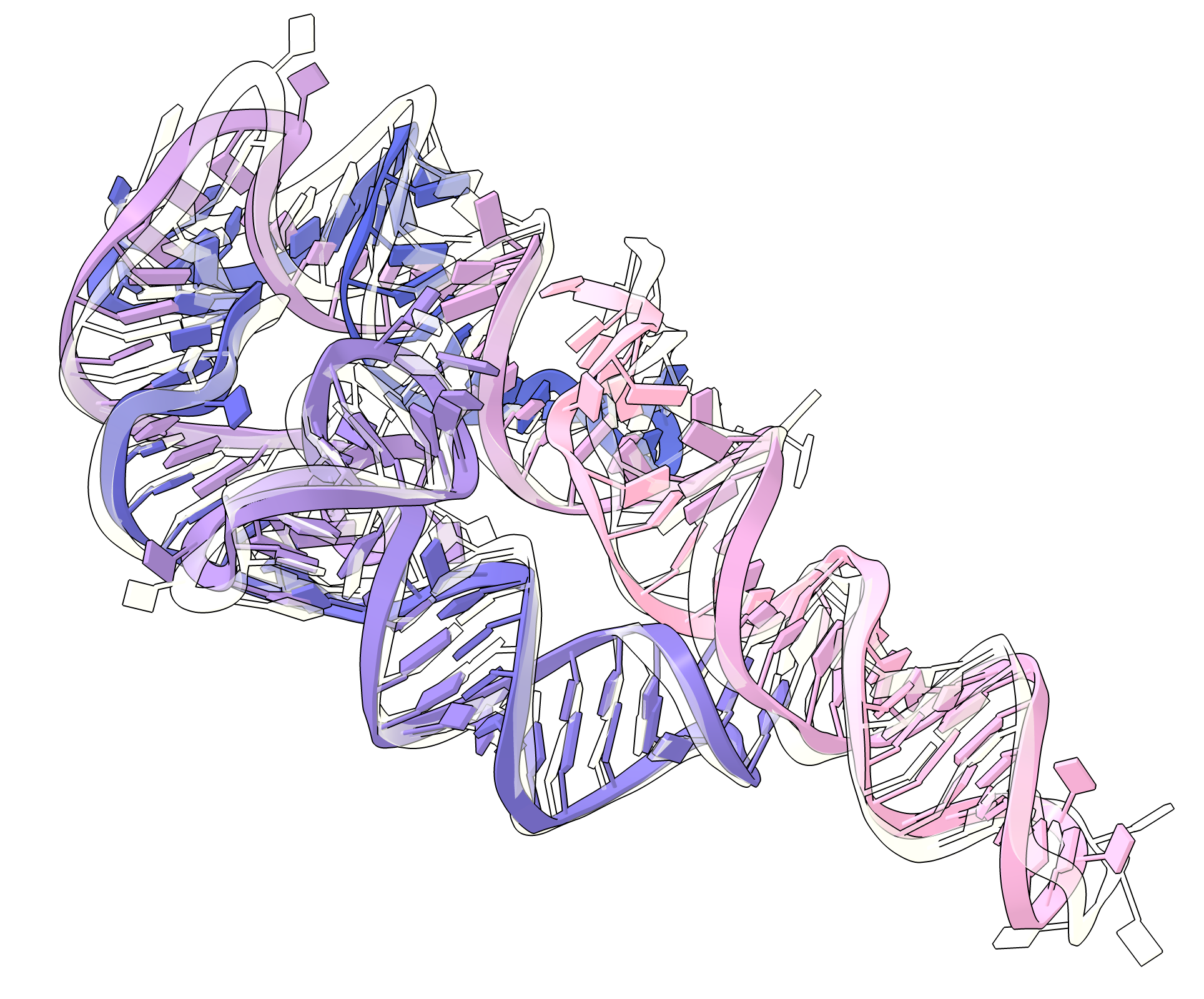}
    {\ttfamily
    \\
      GDT\_TS = 0.776 \\
      TM-score = 0.797\\
      RMSD = 1.902
    }
    \caption*{\textbf{PDB: 2R8S}}
  \end{subfigure}
  \hfill
  \begin{subfigure}[t]{0.32\textwidth}
    \centering
    \includegraphics[width=0.55\textwidth]{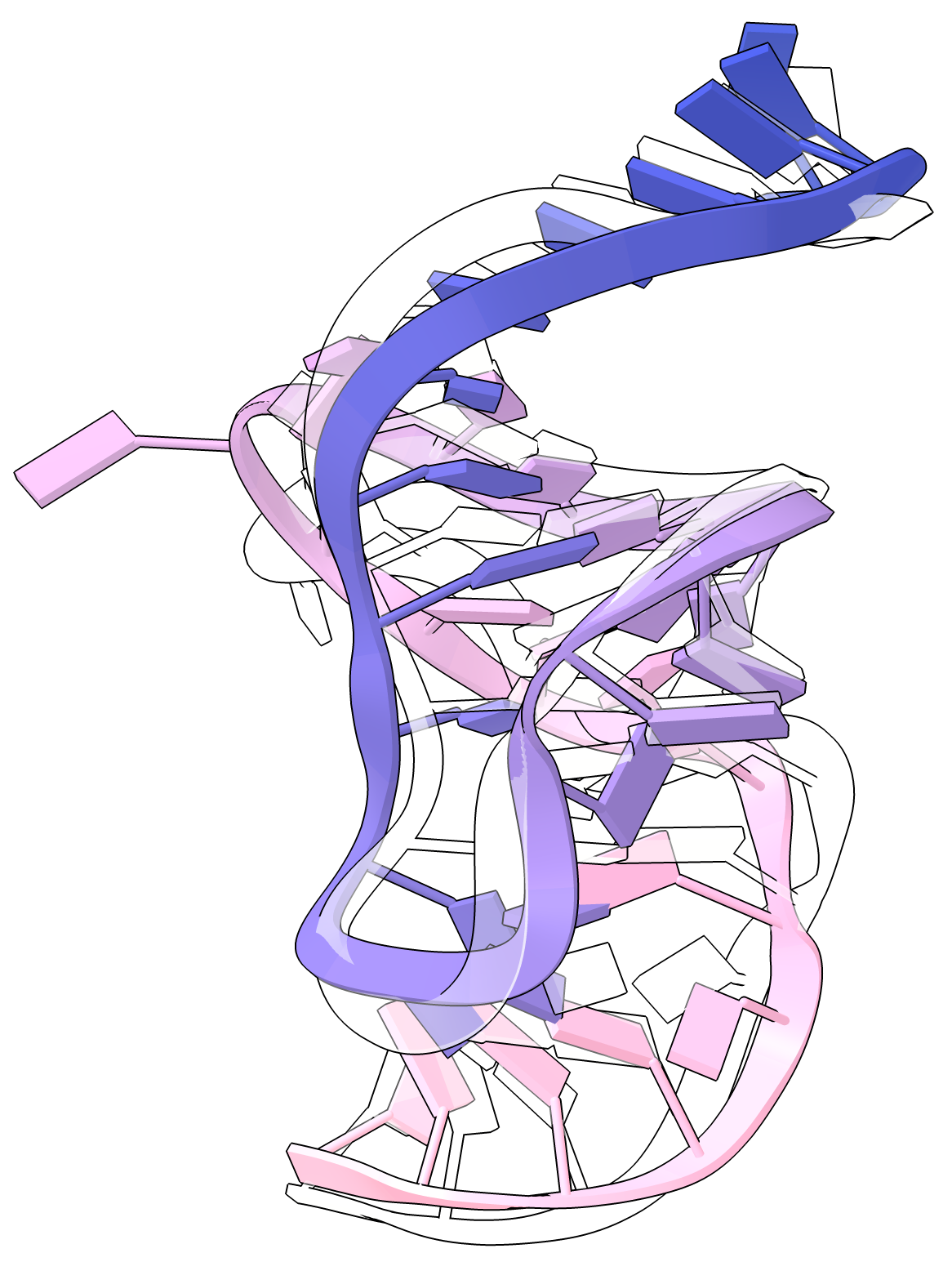}
    {\ttfamily
    \\
      GDT\_TS = 0.643 \\
      TM-score = 0.411\\
      RMSD = 2.819
    }
    \caption*{\textbf{PDB: 1ET4}}
  \end{subfigure}

  \caption{Visualization of designed examples. Structures folded from sequences generated by \our\ are shown in color, while the target structures are shown in semi-transparent yellow.}
  \label{fig:visualization}
\end{figure}

\begin{table}[t]
  \centering
  \renewcommand{\arraystretch}{1.3}
  \caption{Performance comparison of different methods and reward functions on the test set. Results are reported as the mean and standard deviation over 5 independent runs. Percentages in parentheses indicate the proportion of designs that meet the structural designability thresholds (GDT\_TS $\ge 0.5$, TM-score $\ge 0.45$, RMSD $\le 2$\AA).}
  \label{tab:results}
  \resizebox{1.0\linewidth}{!}{
  \begin{tabular}{@{}lccccc@{}}
    \toprule
    \multirow{2}{*}{\textbf{Method}} & \multirow{2}{*}{\textbf{Reward function}} & \multicolumn{3}{c}{\textbf{3D self-consistency metrics}} & \multirow{2}{*}{\textbf{NSR}} \\
    \cmidrule(lr){3-5}
     &  & \textbf{GDT\_TS} $\uparrow$& \textbf{RMSD} $\downarrow$& \textbf{TM-score} $\uparrow$& \\
    \midrule
    \multirow{4}{*}{\our} & $R^{\texttt{tm}}$     & \cellcolor[rgb]{ .702,  .702,  .702}{$\mathbf{0.62 \pm 0.04 (72\%)}$} & $4.31 \pm 0.70 (31\%)$ & \cellcolor[rgb]{ .702,  .702,  .702}{$\mathbf{0.61 \pm 0.03 (72\%)}$} & $0.45\pm0.04$ \\
          & $R^{\texttt{gdt}}$      & $0.58 \pm 0.05 (63\%)$ & $4.60 \pm 0.55 (30\%)$ & $0.56 \pm 0.05 (64\%)$ & $0.47\pm0.06$ \\
          & $R^{\texttt{rmsd}}$       & $0.56 \pm 0.03 (60\%)$ & $4.09 \pm 0.40 (31\%)$ & $0.55 \pm 0.06 (64\%)$ & $0.49\pm0.05$ \\
          & $R^{\texttt{gdt\_rmsd}}$ & \cellcolor[rgb]{ .702,  .702,  .702}{$\mathbf{0.62 \pm 0.02 (72\%)}$} & \cellcolor[rgb]{ .702,  .702,  .702}{$\mathbf{3.35 \pm 0.44 (33\%)}$} & $0.56 \pm 0.03 (68\%)$ & $0.47\pm0.05$ \\
    \midrule
    \texttt{RIDE}     & --            & $0.33 \pm 0.03 (31\%)$ & $10.36 \pm 0.19 (8\%)$ & $0.33 \pm 0.03 (36\%)$ & \cellcolor[rgb]{ .702,  .702,  .702}{$\mathbf{0.61 \pm 0.02}$} \\
    \midrule
    \texttt{RIDE} (Best-of-N)     & --            & $0.42 \pm 0.06 (46\%)$ & $8.63 \pm 0.90 (15\%)$ & $0.39 \pm 0.05 (45\%)$ & $0.50 \pm 0.06$ \\
    \midrule
    RiboDiffusion     & --            & $0.13 \pm 0.02 (5\%)$ & $15.68 \pm 0.26 (1\%)$ & $0.14 \pm 0.03 (6\%)$ & $0.56 \pm 0.04$ \\
    \midrule
    gRNAde & --            & $0.28 \pm 0.06 (27\%)$ & $10.89 \pm 0.61 (3\%)$ & $0.30 \pm 0.05 (28\%)$ & $0.55 \pm 0.03$ \\
    \bottomrule
  \end{tabular}
  }
\end{table}

\begin{table}[t]
\centering
\renewcommand{\arraystretch}{1.3}
\caption{Cross-predictor validation using the independent AlphaFold3 oracle.}
\label{tab:af3}
\resizebox{1.0\linewidth}{!}{
\begin{tabular}{l c c c c}
\toprule
\textbf{Model} & \textbf{Reward function} & \textbf{GDT\_TS} $\uparrow$ & \textbf{RMSD} $\downarrow$ & \textbf{TM-score} $\uparrow$ \\
\midrule
\our\ (AlphaFold3) & $R^{\texttt{tm}}$  & $0.56 \pm 0.07(64\%)$ & $4.69 \pm 0.79(30\%)$ & $0.56 \pm 0.06(65\%)$ \\
\our\ (AlphaFold3) & $R^{\texttt{gdt\_rmsd}}$  & \cellcolor[rgb]{ .702,  .702,  .702}{$\mathbf{0.57 \pm 0.06(65\%)}$} & \cellcolor[rgb]{ .702,  .702,  .702}{$\mathbf{3.80 \pm 0.61(32\%)}$} & \cellcolor[rgb]{ .702,  .702,  .702}{$\mathbf{0.56 \pm 0.05(67\%)}$} \\
gRNAde (AlphaFold3) & -- & $0.26 \pm 0.04(26\%)$ & $9.96 \pm 0.77(3\%)$ & $0.28 \pm 0.04(26\%)$ \\
\bottomrule
\end{tabular}
}
\end{table}

\subsection{Reinforcement learning fine-tuning}

After pre-training, we further fine-tune \texttt{RIDE} using reinforcement learning, denoted as \our.

\paragraph{\our\ achieves substantial improvements in 3D structural similarity}
Figure~\ref{fig:rl_results}A shows results on $14$ RNA structures of interest identified by~\citet{das2010atomic}, all of which are included in the test set. \our\ delivers significant improvements on the vast majority of these structures, with more than $100\%$ improvement on half of them. Table~\ref{tab:results} summarizes the complete results across the test set. Across all three 3D self-consistency metrics, \our\ achieves more than $100\%$ improvement relative to the best-performing baseline. Using GDT\_TS as the evaluation criterion, $72\%$ of \our\ designs exceed the threshold of $0.5$, compared to only $27\%$ for gRNAde. In terms of RMSD, $33\%$ of \our\ sequences achieve $\le 2$\AA, compared to only $3\%$ for gRNAde. Figure~\ref{fig:visualization} illustrates three representative examples of designed structures. 

\paragraph{\our\ can discover high-quality designs different from native sequences}
Figure~\ref{fig:rl_results}B shows the change in NSR before and after RL fine-tuning. The color of each point denotes the GDT\_TS value after fine-tuning. We observe that, in most cases, the NSR of \our\ designs is lower than that of \texttt{RIDE}, yet these sequences achieve higher GDT\_TS. 

\paragraph{\our\ generalizes across different oracles}
To further evaluate the generalization capability of \our\ models, we design a variant using AlphaFold3~
\citep{abramson2024accurate} to replace the original RhoFold. From the results in Table~\ref{tab:af3}, we find that \our\ maintains its superiority under this cross-oracle evaluation. In particular, there are minor performance differences when using different oracles: with the $R^{\texttt{gdt\_rmsd}}$ reward, our GDT\_TS score is $0.57$ with AlphaFold3, which is more than doubled (+$119\%$) the score of the baseline ($0.26$). These results demonstrate that \our\ captures generalizable principles of RNA design rather than overfitting a single predictor.

\paragraph{Impact of reward functions}
Table~\ref{tab:results} reports the impact of the four reward functions introduced in Section~\ref{subsec:rl_finetune}. The TM-score-based reward $R^{\texttt{tm}}$ performs strongly on both TM-score and GDT\_TS, though it is slightly less effective on RMSD. In contrast, $R^{\texttt{rmsd}}$ and $R^{\texttt{gdt}}$ yield weaker results when used individually. Notably, the composite reward $R^{\texttt{gdt\_rmsd}}$, which combines GDT\_TS and RMSD, achieves the most balanced performance overall. We attribute this to the complementary nature of the two metrics, which jointly capture both local atomic accuracy and global fold similarity. Appendix~\ref{sec:rwd_ablation} provides additional evaluation results for all four reward functions.



%

\section{Conclusion}
\label{sec:conclusion}

We propose a novel two-stage framework for 3D RNA inverse design that directly optimizes structural similarity. In the first stage, we develop \texttt{RIDE}, a conditional diffusion model trained via supervised learning that surpasses SOTA methods in native sequence recovery. In the second stage, we develop \texttt{RIDER}, which significantly improves the similarity between designed and target structures through reinforcement learning. Future work may extend this framework with multi-objective rewards and explore experimental validation of the generated sequences.

\section*{Acknowledgment}
This work was supported by the UKRI Future Leaders Fellowship under Grant MR/S017062/1 and MR/X011135/1; in part by National Natural Science Foundation of China under Grant 62373375, U2341216, 62376056 and 62076056; in part by the Science and Technology Innovation Program of Hunan Province under Grant 2024RC1011 and the Natural Science Foundation of Hunan Province under Grant 2025JJ10007; in part by the Isambard-AI, Royal Society Faraday Discovery Fellowship (FDF/S2/251014), BBSRC Transformative Research Technologies (UKRI1875), Royal Society International Exchanges Award (IES/R3/243136), Kan Tong Po Fellowship (KTP/R1/231017); and the Amazon Research Award and Alan Turing Fellowship.

\bibliographystyle{plainnat}
\bibliography{references/references}

\appendix
\newpage

\appendix

\section{Algorithm Pseudocode}
\label{sec:appendix_algorithm}

This section provides detailed pseudocode for the key algorithmic components of our proposed method. Specifically, we include three main procedures:

\begin{itemize}
\item \textbf{Algorithm~\ref{alg:pretrain}} describes the supervised pre-training stage of the conditional diffusion model (\texttt{RIDE}), where the model learns to predict noise added to clean RNA sequences given a target 3D structure.
\item \textbf{Algorithm~\ref{alg:sample}} presents the DDIM-based sampling procedure for generating RNA sequences conditioned on a target structure using the pre-trained model. This forms the basis of both inference and experience collection in the RL stage.
\item \textbf{Algorithm~\ref{alg:rl}} outlines the reinforcement learning fine-tuning stage (\texttt{RIDER}), where a policy gradient method is used to optimize the diffusion model toward improved structural fidelity based on 3D evaluation metrics (GDT\_TS, TM-score, RMSD).
\end{itemize}

Together, these algorithms define the full learning pipeline of our framework.

\begin{algorithm}[H]
\caption{\texttt{RIDE}: Pre-training of Conditional Diffusion Model}
\label{alg:pretrain}
\begin{algorithmic}[1]
\REQUIRE Training dataset $\mathcal{D} = \{(y_0^{(i)}, X_{\text{target}}^{(i)})\}$, number of diffusion steps $T_{\text{diff}}$, model parameters $\theta$
\REQUIRE Structure encoder $E_{\text{gnn}}$, GVP-GNN noise prediction network $\epsilon_\theta$  
\ENSURE Pre-trained model parameters $\theta$

\STATE Initialize model parameters $\theta$ (for $\epsilon_\theta$)
\FOR{each training iteration}
    \STATE Sample a minibatch $(y_0, X_{\text{target}})$ from $\mathcal{D}$
    \STATE Compute structure condition $\mathcal{C} = E_{\text{gnn}}(X_{\text{target}})$
    \STATE Sample timestep $t \sim \text{Uniform}(\{1, \dots, T_{\text{diff}}\})$
    \STATE Sample standard Gaussian noise $\varepsilon \sim \mathcal{N}(0, I)$
    \STATE Generate noisy sequence $y_t = \alpha_t y_0 + \sigma_t \varepsilon$ using the forward process
    \STATE Predict noise using the model: $\hat{\varepsilon} = \epsilon_\theta(y_t, t, \mathcal{C})$
    \STATE Compute pre-training loss: $L_{\text{pretrain}} = \|\varepsilon - \hat{\varepsilon}\|^2$
    \STATE Compute gradients $\nabla_\theta L_{\text{pretrain}}$
    \STATE Update parameters: $\theta \leftarrow \theta - \eta \nabla_\theta L_{\text{pretrain}}$
\ENDFOR
\STATE \RETURN Trained model parameters $\theta$
\end{algorithmic}
\end{algorithm}

\begin{algorithm}[H]
\caption{Sequence Sampling via DDIM}
\label{alg:sample}
\begin{algorithmic}[1]
\REQUIRE Target structure $X_{\text{target}}$, trained or fine-tuned model parameters $\theta$ (for $\epsilon_\theta$)  
\REQUIRE Structure encoder $E_{\text{gnn}}$, GVP-GNN noise prediction network $\epsilon_\theta$  
\REQUIRE Total number of sampling steps $N_{\text{steps}}$ 
\ENSURE Designed RNA sequence $\hat{y}_0$

\STATE Compute structure condition $\mathcal{C} = E_{\text{gnn}}(X_{\text{target}})$
\STATE Sample initial noise vector $y_{N_{\text{steps}}} \sim \mathcal{N}(0, I)$ 
\FOR{$k = N_{\text{steps}}$ \textbf{down to} $1$}
    \STATE Determine current and previous time steps $t_k$ and $t_{k-1}$
    \STATE Predict noise: $\hat{\varepsilon}_k = \epsilon_\theta(y_{t_k}, t_k, \mathcal{C})$
    \STATE Compute $y_{t_{k-1}}$ using DDIM update rule with $y_{t_k}$ and $\hat{\varepsilon}_k$
\ENDFOR
\STATE Set $\hat{y}_0 \leftarrow y_0$ 
\STATE Obtain discrete sequence: $\hat{y}_{0,\text{discrete}} = \text{argmax}(\hat{y}_0)$
\STATE \RETURN $\hat{y}_{0,\text{discrete}}$
\end{algorithmic}
\end{algorithm}

\begin{algorithm}[H]
\caption{\texttt{RIDER}: Reinforcement Learning Fine-Tuning}
\label{alg:rl}
\begin{algorithmic}[1]
\REQUIRE Pretrained policy network $\pi_\theta$; target structure dataset $\mathcal{D}_{\text{target}}$; structure encoder $E_{\text{gnn}}$; RNA structure predictor $F_{\text{pred}}$; reward function $R$
\REQUIRE Clipping coefficient $\epsilon_{\text{clip}}$; learning rate $\eta$; number of RL epochs $N$; number of trajectories per epoch $M_{\text{traj}}$; number of policy update steps per epoch $K_{\text{update}}$
\ENSURE Fine-tuned policy parameters $\theta$

\FOR{$epoch = 1$ \textbf{to} $N$}
    \STATE $\theta_{\text{old}} \leftarrow \theta$
    \STATE Initialize experience buffer $\mathcal{B}$
    \FOR{$m = 1$ \textbf{to} $M_{\text{traj}}$}
        \STATE Sample target structure $X_{\text{target}}$ from $\mathcal{D}_{\text{target}}$
        \STATE Compute structure condition $\mathcal{C} = E_{\text{gnn}}(X_{\text{target}})$
        \STATE $(\hat{y}_0, \text{trajectory\_data}_m) \leftarrow \text{sampling}$
        \STATE $X_{\text{pred}} = F_{\text{pred}}(\hat{y}_0)$
        \STATE $R_{\text{traj}}^{(m)} = R(X_{\text{pred}}, X_{\text{target}})$
        \STATE Store $(\text{trajectory\_data}_m, R_{\text{traj}}^{(m)})$ in $\mathcal{B}$
    \ENDFOR
    \STATE Compute baseline $b$
    \FOR{$n = 1$ \textbf{to} $K_{\text{update}}$}
        \FOR{each $(\text{trajectory\_data}, R_{\text{traj}})$ in $\mathcal{B}$}
            \STATE Compute advantage: $A = R_{\text{traj}} - b$
            \FOR{each $(s_k, a_k, \log \pi_{\theta_{\text{old}}}(a_k|s_k))$ in trajectory}
                \STATE Compute $\log \pi_\theta(a_k|s_k)$
                \STATE Compute importance ratio $r_k(\theta) = \exp(\log \pi_\theta(a_k|s_k) - \log \pi_{\theta_{\text{old}}}(a_k|s_k))$
                \STATE $L_{\text{traj}} \leftarrow L_{\text{traj}} + \min(r_k(\theta) A, \text{clip}(r_k(\theta), 1 - \epsilon_{\text{clip}}, 1 + \epsilon_{\text{clip}}) A)$
            \ENDFOR
        \ENDFOR
        \STATE $L^{\text{RL}} \leftarrow L_{\text{traj}}/ M_{\text{traj}}$
        \STATE Compute gradient $\nabla_\theta L^{\text{RL}}$
        \STATE Update parameters: $\theta \leftarrow \theta - \eta \nabla_\theta L^{\text{RL}}$
    \ENDFOR
\ENDFOR
\STATE \RETURN Fine-tuned parameters $\theta$
\end{algorithmic}
\end{algorithm}

\section{Model architecture and training hyperparameters}
\label{sec:appendix_hyperparameters}

\begin{figure}[t]
\centering
\includegraphics[width=1\linewidth]{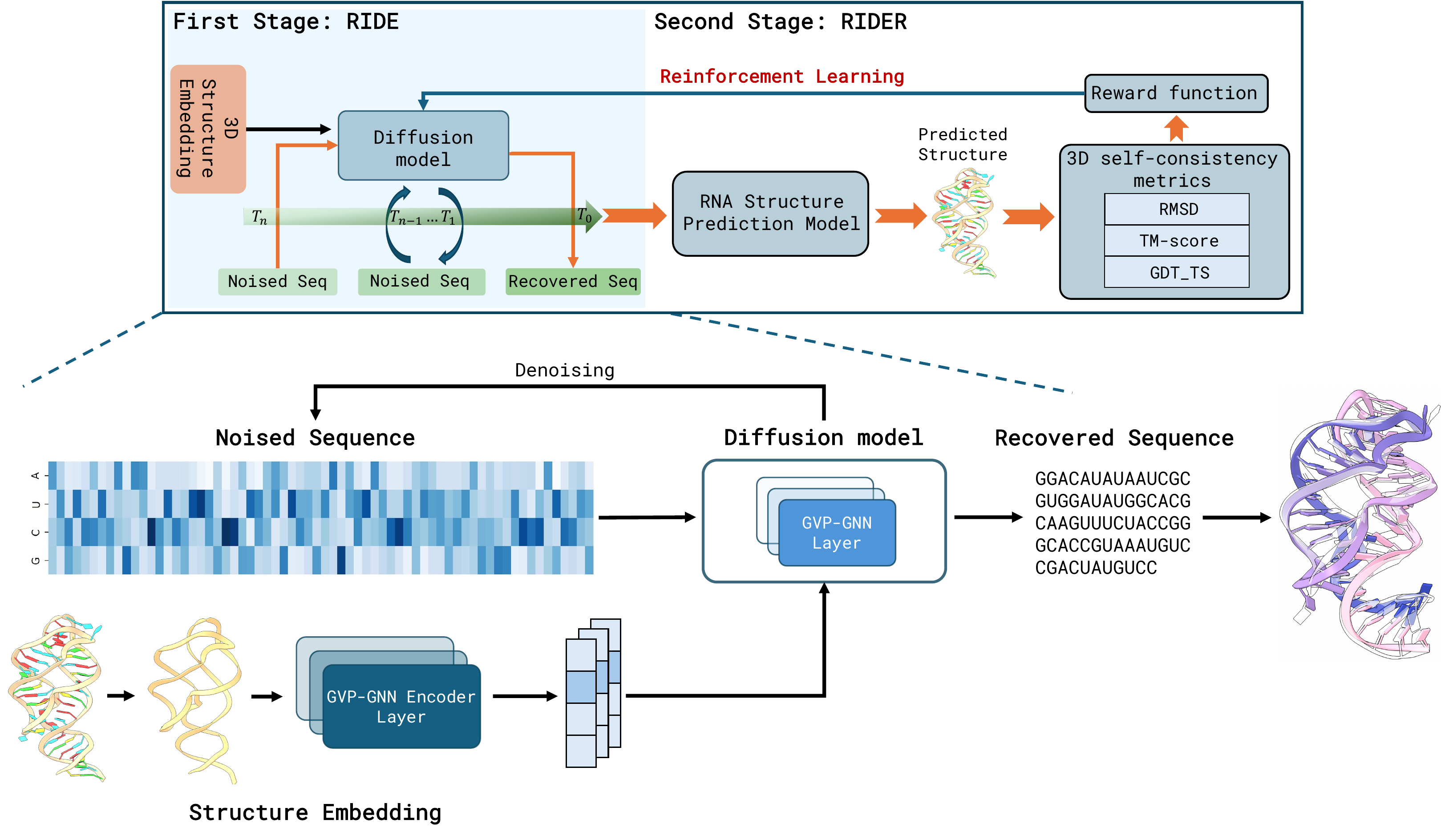}
\caption{The proposed framework consists of two stages: \texttt{RIDE} (pre-training via supervised learning) and \texttt{RIDER} (fine-tuning via reinforcement learning). A multi-layer GVP-GNN encoder-decoder backbone is used to iteratively denoise RNA sequences conditioned on the target 3D structure. Structural fidelity is evaluated using a separate RNA structure predictor and fed back through a reward function.}
\label{fig:overview_two_stage}
\end{figure}

This section provides detailed descriptions of the \texttt{RIDE} architecture, pre-training and fine-tuning hyperparameters, as well as the computational setup used during training. Figure~\ref{fig:overview_two_stage} illustrates the overall architecture and learning pipeline of our proposed method. The framework consists of two stages: (1) \texttt{RIDE}, a conditional diffusion model that is pre-trained to generate RNA sequences from target 3D structures, and (2) \texttt{RIDER}, a reinforcement learning fine-tuning stage that optimizes the diffusion model for structural fidelity using 3D self-consistency metrics (GDT\_TS, TM-score, RMSD). The model takes as input a 3D structure embedding and iteratively denoises a sequence initialized from Gaussian noise, guided by a multi-layer GVP-GNN encoder-decoder architecture. The predicted sequences are evaluated via a structure prediction model, and reward signals are computed to fine-tune the policy.

\subsection{Structure representation}
\label{sec:appendix_representation}

To effectively capture the geometric intricacies of RNA tertiary structures, we first convert the input backbone into a graph representation, which is then processed by a Geometric Vector Perceptron Graph Neural Network (GVP-GNN) \citep{jing2021learning} encoder.

\paragraph{Featurization}
We represent each RNA tertiary structure as a geometric graph $\mathcal{G} = (\mathcal{V}, \mathcal{E})$, where $\mathcal{V}$ denotes nucleotides as nodes and $\mathcal{E}$ encodes spatial adjacencies. Each nucleotide $i$ is characterized by the coordinates of three backbone atoms: P, C4', and N1 (for pyrimidines) or N9 (for purines). The node coordinate $\vec{x}_i \in \mathbb{R}^3$ is defined as the centroid of these atoms. To capture local geometry, edges connect each node to its $k=32$ nearest spatial neighbors.

Node features $h_i^{(0)}$ consist of scalar features $s_i \in \mathbb{R}^{d_s^{in}}$ and vector features $\vec{v}_i \in \mathbb{R}^{d_v^{in} \times 3}$. Scalar features include intra-nucleotide distances and angles, while vector features capture backbone orientation, such as the unit vector $\vec{x}_{i+1}-\vec{x}_i$ and vectors from C4' to P and N1/N9. Edge features $e_{ij}^{(0)}$ for an edge between node $i$ and $j$ similarly consist of scalar components $s_{ij} \in \mathbb{R}^{d_{es}^{in}}$ (e.g., 3D distance $||\vec{x}_j - \vec{x}_i||_2$ encoded via radial basis functions, and sequence separation $j-i$ encoded via sinusoidal positional encodings) and vector components $\vec{e}_{ij} \in \mathbb{R}^{d_{ev}^{in} \times 3}$ (e.g., the unit vector $\vec{x}_j - \vec{x}_i$).

\paragraph{Structure encoder}
The featurized graph is processed by a structure encoder composed of $L_E=5$ GVP-GNN layers. Each layer $l$ updates the node representations $(s_i^{(l)}, \vec{v}_i^{(l)})$ by aggregating information from neighboring nodes and edges in an $E(3)$-equivariant manner. The update rule for a node $i$ at layer $l$ can be expressed as:
\begin{align}
    m_{i}^{(l)}, \vec{m}_{i}^{(l)} &= \underset{j \in \mathcal{N}(i)}{\text{AGG}} \left( \phi_{\texttt{msg}}^{(l)} \left( (s_i^{(l-1)}, \vec{v}_i^{(l-1)}), (s_j^{(l-1)}, \vec{v}_j^{(l-1)}), (s_{ij}, \vec{e}_{ij}) \right) \right) \label{eq:encoder_msg}, \\
    s_i^{(l)}, \vec{v}_i^{(l)} &= \phi_{\texttt{update}}^{(l)} \left( (s_i^{(l-1)}, \vec{v}_i^{(l-1)}), (m_i^{(l)}, \vec{m}_i^{(l)}) \right), \label{eq:encoder_update}
\end{align}
where $\mathcal{N}(i)$ denotes the neighbors of node $i$, $\phi_{\texttt{msg}}^{(l)}$ and $\phi_{\texttt{update}}^{(l)}$ are GVP-based message and update functions respectively, and AGG is an aggregation operator (e.g., mean).
The encoder network first embeds the initial node and edge features using GVP layers $W_v$ and $W_e$:
\begin{align}
    (s_i^{(0)}, \vec{v}_i^{(0)}) = W_v((s_i^{in}, \vec{v}_i^{in})), \;\;\;
    (s_{ij}^{(0)}, \vec{e}_{ij}^{(0)}) = W_e((s_{ij}^{in}, \vec{e}_{ij}^{in})).
\end{align}
The final output of the structure encoder is a set of node-level conditional embeddings $h_c = \{(s_i^{(L_E)}, \vec{v}_i^{(L_E)})\}$ for all $i \in \mathcal{V}$. These embeddings, $h_c$, encapsulate the essential 3D structural information and serve as the conditioning context for the diffusion model.

\subsection{\texttt{RIDE}: Conditional Diffusion Model Architecture}
\label{sec:dmgnn_architecture}

The \texttt{RIDE} model is a conditional diffusion network built upon Geometric Vector Perceptron Graph Neural Networks (GVP-GNNs), specifically designed for RNA inverse folding.

\paragraph{GVP-GNN configuration.}
Both the structure encoder and the noise prediction network within the diffusion model comprise $L_E = L_D = 5$ layers of GVP-GNNs. The key architectural parameters and node/edge feature dimensions are summarized in Table~\ref{tab:gnn_features_hyperparams}. A dropout rate of $0.5$ is applied to each GVP-GNN layer to mitigate overfitting. To enhance robustness, Gaussian noise with a standard deviation of $0.1$ is added to node coordinates during training. The model incorporates time-step embeddings as additional conditioning inputs to guide the denoising process. In total, \texttt{RIDE} contains $10,214,213$ trainable parameters.

\begin{table}[h!]
  \caption{\texttt{RIDE} architecture and associated hyperparameters.}
  \label{tab:gnn_features_hyperparams}
  \centering
  \renewcommand{\arraystretch}{1.2}
  \begin{tabular}{lc}
    \toprule
    \textbf{Hyperparameter} & \textbf{Value} \\
    \midrule
    Number of GVP-GNN layers ($L_E$, $L_D$) & 5 \\
    Node input dimensions (scalar, vector) & (15, 4) \\
    Node hidden dimensions (scalar, vector) & (256, 24) \\
    Edge input dimensions (scalar, vector) & (131, 3) \\
    Edge hidden dimensions (scalar, vector) & (128, 4) \\
    Dropout rate & 0.5 \\
    Output dimension & 4 \\
    \midrule
    Number of nearest neighbors per node & 32 \\
    Number of radial basis functions (RBFs) & 32 \\
    Positional encoding dimension for edges & 32 \\
    \bottomrule
  \end{tabular}
\end{table}

\paragraph{Diffusion model pre-training.}
\texttt{RIDE} is pre-trained via noise prediction using a Mean Squared Error (MSE) loss. The forward noising process follows a Variance-Preserving (VP) Stochastic Differential Equation (SDE). Key hyperparameters for the SDE and training schedule are summarized in Table~\ref{tab:dm_gnn_pretrain_hyperparams}. The initial learning rate is set to $3 \times 10^{-4}$ and is decayed by a factor of $0.9$ using a \texttt{ReduceLROnPlateau} scheduler when the validation performance (measured by NSR) fails to improve for 5 consecutive epochs.

\begin{table}[h!]
  \caption{Training hyperparameters for \texttt{RIDE} pre-training.}
  \label{tab:dm_gnn_pretrain_hyperparams}
  \centering
  \renewcommand{\arraystretch}{1.2}
  \begin{tabular}{lc}
    \toprule
    \textbf{Hyperparameter} & \textbf{Value} \\
    \midrule
    SDE schedule & Linear \\
    Initial noise scale $\beta_0$ & 0.1 \\
    Final noise scale $\beta_1$ & 20.0 \\
    Total diffusion time $T$ & 1.0 \\
    Minimum time step $\epsilon_{\texttt{time}}$ & 0.001 \\
    \midrule
    Number of epochs & 150 \\
    Initial learning rate & $3 \times 10^{-4}$ \\
    Learning rate scheduler & \texttt{ReduceLROnPlateau} \\
    Scheduler decay factor & 0.9 \\
    Scheduler patience & 5 epochs \\
    Optimizer & \texttt{Adam} \\
    Activation function & \texttt{SiLU} \\
    Max nodes per batch & 3000 \\
    Max nodes per sample & 5000 \\
    \midrule
    DDIM denoising steps $N_{\texttt{steps}}$ & 50 \\
    Sampling temperature & 0.1 \\
    \bottomrule
  \end{tabular}
\end{table}

\subsection{Details of \texttt{RIDER}}
\label{sec:dmrl_details}

Following pre-training, the \texttt{RIDE} model is fine-tuned using an advantage-based policy gradient algorithm, referred to as \texttt{RIDER}.

\paragraph{RL Algorithm Settings.}
Reinforcement learning fine-tuning is performed using the Adam optimizer without any learning rate scheduling. Key hyperparameters for the training loop, policy updates, and experience collection are summarized in Table~\ref{tab:dm_rl_hyperparams}.

\begin{table}[h!]
  \caption{Training hyperparameters for \texttt{RIDER} fine-tuning.}
  \label{tab:dm_rl_hyperparams}
  \centering
  \renewcommand{\arraystretch}{1.2}
  \begin{tabular}{lc}
    \toprule
    \textbf{Hyperparameter} & \textbf{Value} \\
    \midrule
    Learning rate & $5 \times 10^{-5}$ \\
    Optimizer & Adam \\
    Number of RL training epochs & 80 \\
    Policy updates per epoch & 2 \\
    Gradient accumulation steps & 60 \\
    Clip range $\epsilon_{\texttt{clip}}$ & 0.5 \\
    Max gradient norm & 1.0 \\
    Batch size & 60 \\
    DDIM denoising steps during RL & 30 \\
    \bottomrule
  \end{tabular}
\end{table}

\paragraph{Reward Function Scaling Factors.}
The reward functions used during RL fine-tuning based on three structural similarity metrics. The scaling coefficients for these components, as defined in Section~\ref{subsec:rl_finetune}, are summarized in Table~\ref{tab:reward_scaling_factors}.

\begin{table}[h!]
  \caption{Scaling factors for reward function components.}
  \label{tab:reward_scaling_factors}
  \centering
  \renewcommand{\arraystretch}{1.2}
  \begin{tabular}{lc}
    \toprule
    \textbf{Component} & \textbf{Scaling Factor} \\
    \midrule
    GDT\_TS weight ($w_{\texttt{gdt\_scale}}$) & 5 \\
    TM-score weight ($w_{\texttt{tm\_scale}}$) & 5 \\
    RMSD weight ($w_{\texttt{rmsd\_scale}}$) & 0.5 \\
    GDT\_TS weight for bonus reward ($w_{\texttt{bonus\_gdt}}$) & 100 \\
    RMSD weight for bonus reward ($w_{\texttt{bonus\_rmsd}}$) & 20 \\
    \bottomrule
  \end{tabular}
\end{table}

\subsection{Training Hardware}

All model pre-training and reinforcement learning fine-tuning experiments were conducted on servers equipped with NVIDIA RTX 4090 GPUs and Intel Core i9-13900K CPUs. With automatic mixed precision (AMP) enabled, training requires approximately 16 GB of GPU memory; without AMP, the memory usage increases to around 23 GB. RL fine-tuning requires approximately 8 hours on a single RTX 4090 GPU, while the critical inference cost remains unchanged and highly efficient.

\begin{figure*}[htb]
\vskip 0.1in
\center
    \includegraphics[width=0.7\linewidth]{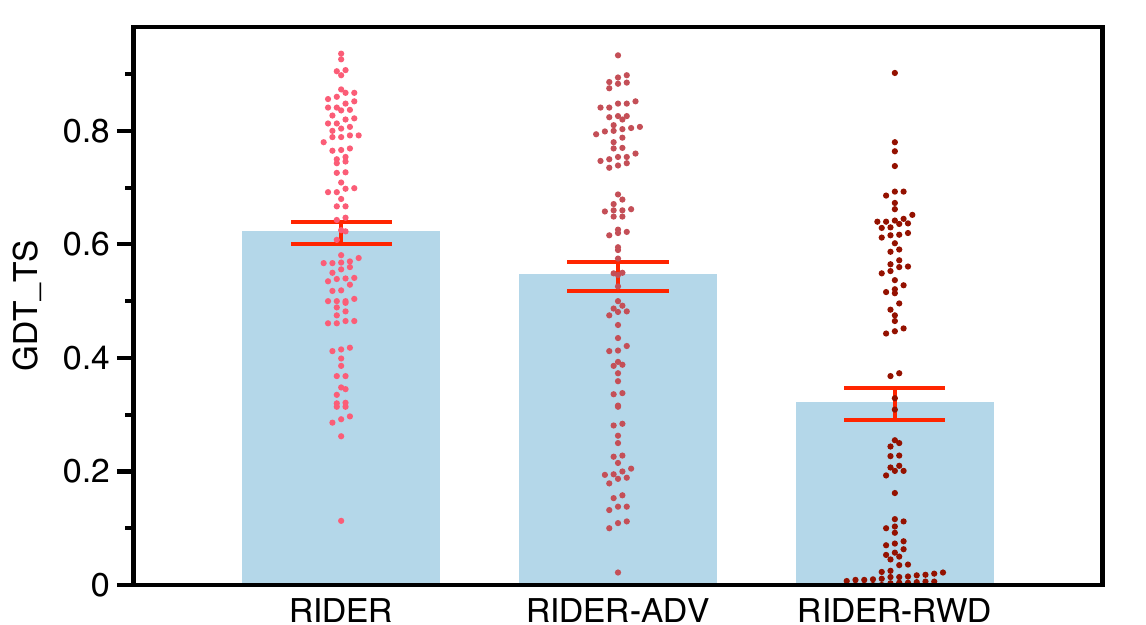}
    \caption{Comparison of \texttt{RIDER}, \texttt{RIDER-RWD}, and \texttt{RIDER-ADV} on the GDT\_TS metric across the test set. Bar heights represent the mean, and red error bars indicate the Standard Error of the Mean (SEM).}
    \label{fig:ADV_gdt}
\vskip -0.1in
\end{figure*}

\begin{figure*}[t!]
\vskip 0.1in
\center
    \includegraphics[width=0.7\linewidth]{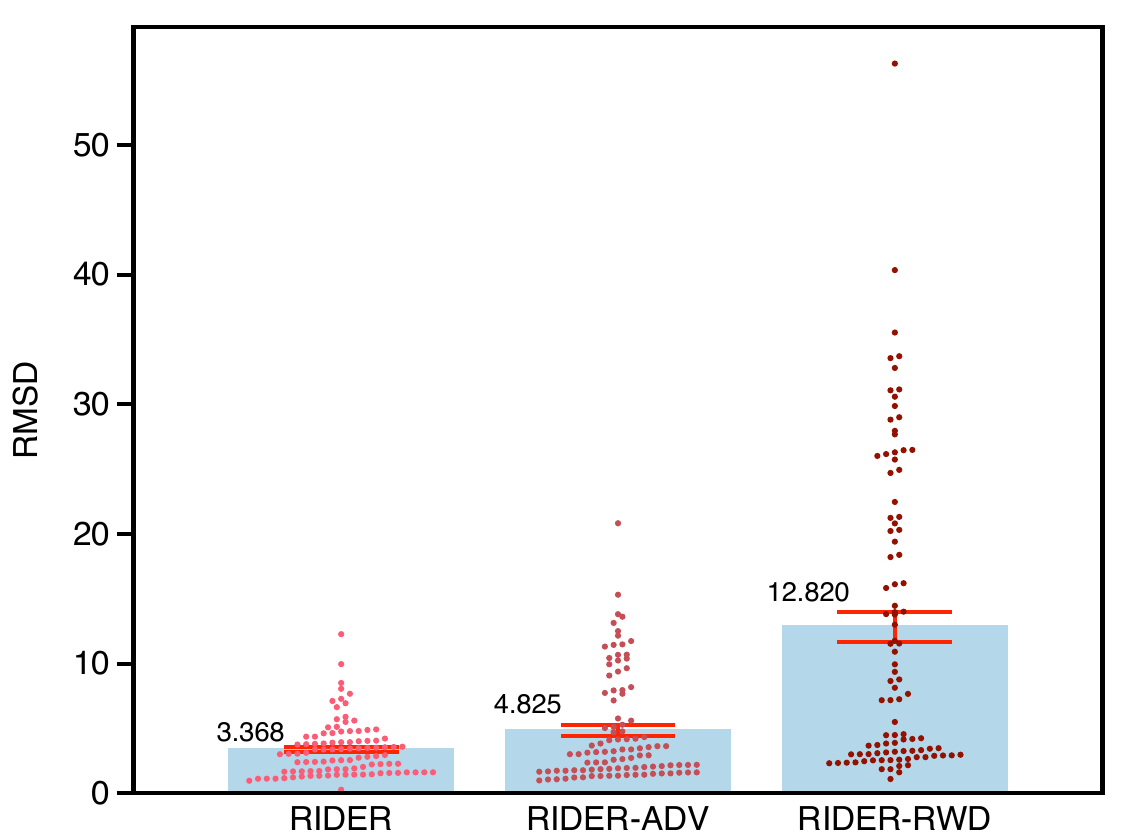}
    \caption{Comparison of \texttt{RIDER}, \texttt{RIDER-RWD}, and \texttt{RIDER-ADV} on the RMSD metric across the test set. Bar heights represent the mean, and red error bars indicate the SEM.}
    \label{fig:ADV_rmsd}
\vskip -0.1in
\end{figure*}

\begin{figure*}[t!]
\vskip 0.1in
\center
    \includegraphics[width=0.7\linewidth]{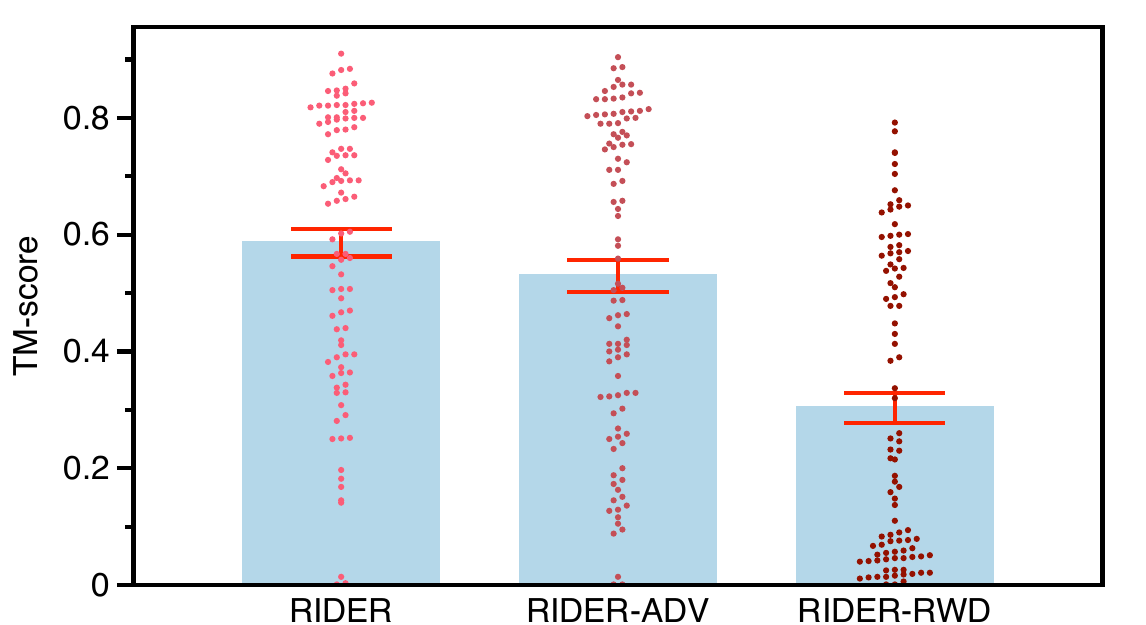}
    \caption{Comparison of \texttt{RIDER}, \texttt{RIDER-RWD}, and \texttt{RIDER-ADV} on the TM-score metric across the test set. Bar heights represent the mean, and red error bars indicate the SEM.}
    \label{fig:ADV_tm}
\vskip -0.1in
\end{figure*}

\section{3D Structure Self‐Consistency Metrics}
\label{sec:3d_metrics}

To evaluate how accurately a designed RNA sequence folds back into its intended 3D backbone structure, we adopt three complementary global metrics: Root Mean Square Deviation (RMSD), Global Distance Test Total Score (GDT\_TS), and Template Modeling Score (TM-score). These metrics capture distinct aspects of structural similarity and are defined as follows.

\subsection{Root‐Mean‐Square Deviation (RMSD)}

RMSD quantifies the average atomic displacement between two structures after optimal superposition. Given two sets of $N$ equivalent atoms with coordinates $\{\mathbf{x}_i\}$ (predicted) and $\{\mathbf{y}_i\}$ (reference), RMSD is defined as:
\begin{equation}
  \text{RMSD} 
    = \sqrt{\frac{1}{N}\sum_{i=1}^N \bigl\|\mathbf{x}_i - \mathbf{y}_i\bigr\|^2}\,.
\end{equation}

Lower RMSD values indicate closer structural alignment. High-accuracy models typically achieve $\text{RMSD} < 2$\AA, while poorly modeled regions may exceed 10\AA.

\subsection{Global Distance Test Total Score (GDT\_TS)}

GDT\_TS quantifies the fraction of residues that can be superimposed within multiple distance cutoffs. For a reference structure of length $N$, let $N_{d_k}$ denote the number of C$\alpha$ atom pairs within a distance threshold $d_k \in \{1, 2, 4, 8\}$\AA, computed under independently optimized superpositions for each $d_k$. GDT\_TS is then defined as:
\begin{equation}
  \text{GDT\_TS} 
    = \frac{1}{4N} \sum_{k \in \{1,2,4,8\}} N_{d_k} \times 100\%.
\end{equation}

By focusing only on well-aligned residues ($d_k \le 8$\AA), GDT\_TS is robust to local distortions and emphasizes the proportion of accurately modeled structure.

\begin{figure*}[t!]
\center
    \includegraphics[width=1.0\linewidth]{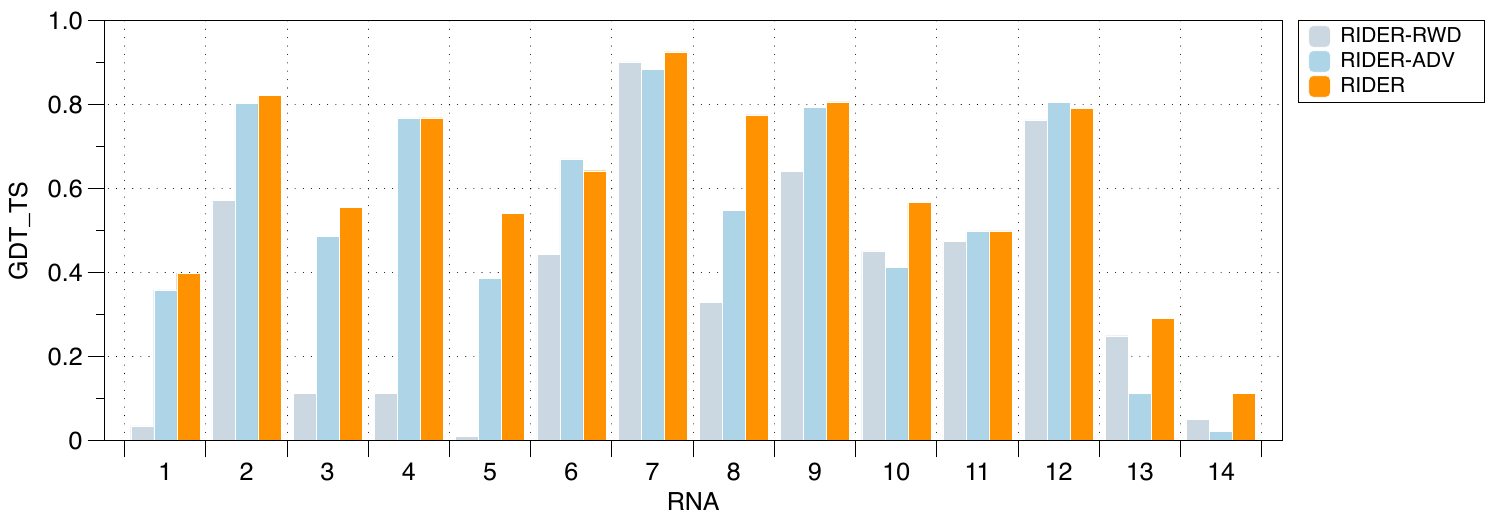}
    \caption{GDT\_TS scores of \texttt{RIDER}, \texttt{RIDER-RWD}, and \texttt{RIDER-ADV} on 14 target RNA structures.}
    \label{fig:ADV_14_gdt}
\end{figure*}

\begin{figure*}[t!]
\center
    \includegraphics[width=1.0\linewidth]{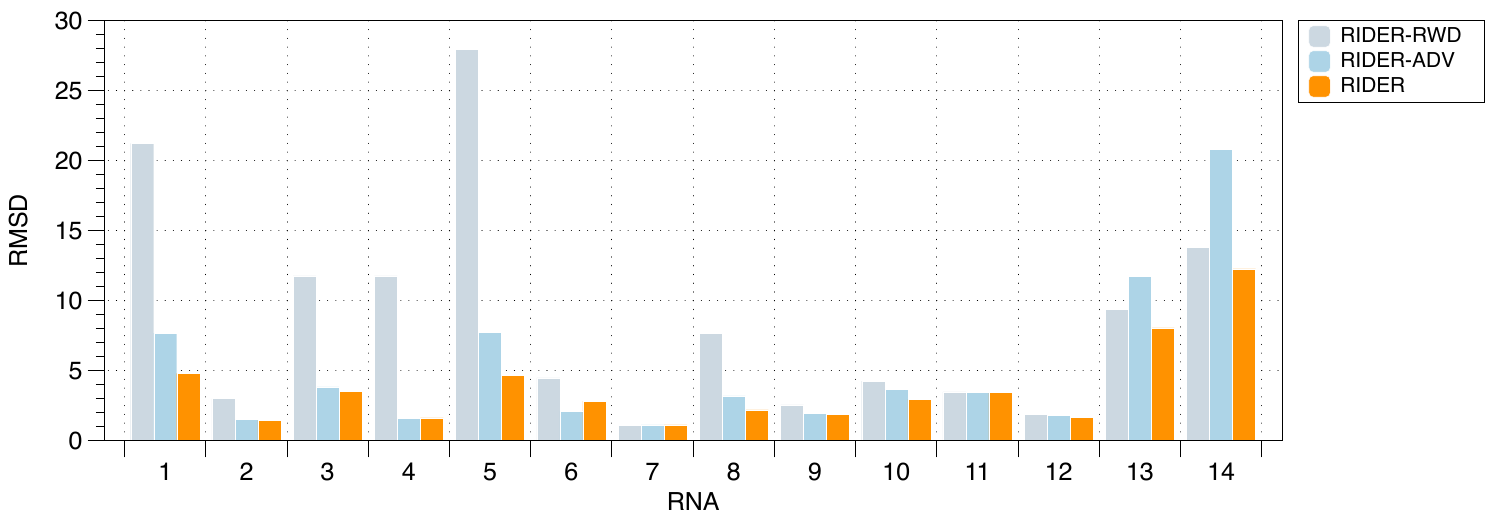}
    \caption{RMSD scores of \texttt{RIDER}, \texttt{RIDER-RWD}, and \texttt{RIDER-ADV} on 14 target RNA structures. Lower values indicate better structural alignment.}
    \label{fig:ADV_14_rmsd}
\end{figure*}

\begin{figure*}[t!]
\center
    \includegraphics[width=1.0\linewidth]{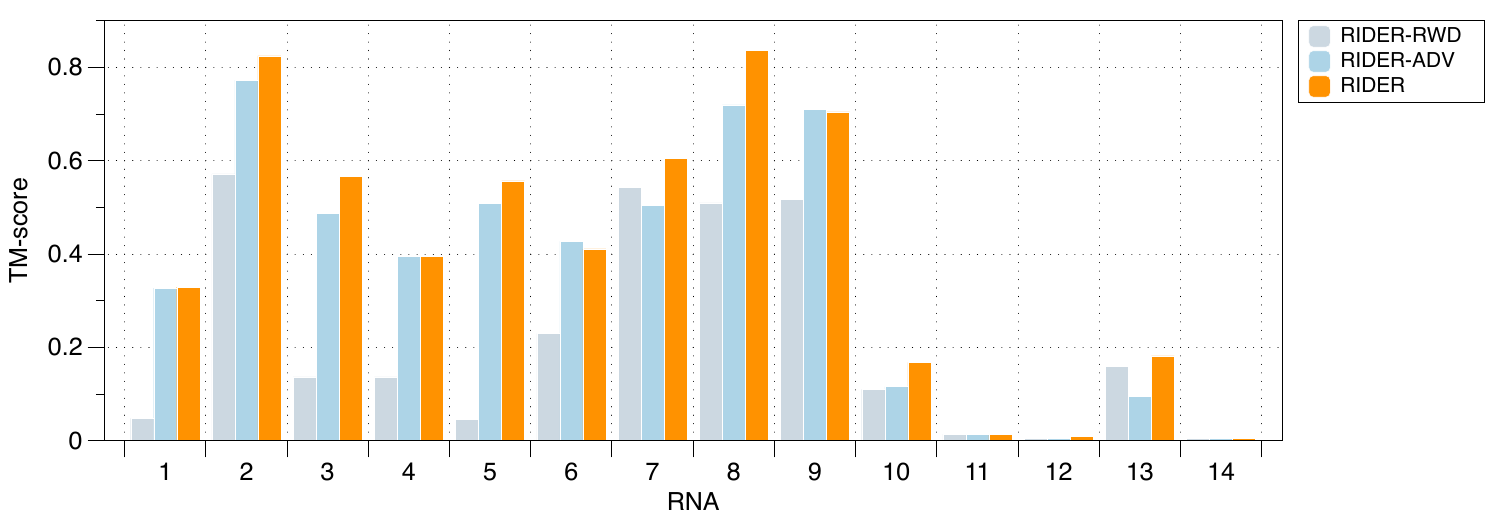}
    \caption{TM-score values of \texttt{RIDER}, \texttt{RIDER-RWD}, and \texttt{RIDER-ADV} on 14 target RNA structures. Higher values indicate better structural similarity.}
    \label{fig:ADV_14_tm}
\end{figure*}

\subsection{Template Modeling Score (TM-score)}

TM-score evaluates global fold similarity using a single optimized superposition with a length-dependent normalization term. Given $L_N$ residues in the reference structure and $L_T$ aligned residue pairs with pairwise distances $d_i$, the TM-score is defined as:
\begin{equation}
  \text{TM-score}
    = \max \left[
      \frac{1}{L_N}
      \sum_{i=1}^{L_T}
        \frac{1}{1 + \bigl(\tfrac{d_i}{d_0(L_N)}\bigr)^2}
        \right],
\end{equation}
where the normalization scale $d_0$ is a function of $L_N$:
\begin{equation}
  d_0(L_N) = 1.24\sqrt[3]{L_N - 15} - 1.8.
\end{equation}

This length correction makes TM-score more stringent for short structures (smaller $d_0$) and more permissive for longer ones, enabling fair comparison across proteins or RNAs of varying lengths.

\subsection{Comparison of Metrics}

RMSD provides a direct measure of average atomic deviation and is highly sensitive to outliers. GDT\_TS quantifies the percentage of residues that fall within multiple distance thresholds, emphasizing well-modeled regions. TM-score yields a length-normalized global similarity score, which is less sensitive to local deviations and more suitable for comparing structures of different sizes. Together, these three metrics offer a comprehensive evaluation framework for assessing 3D structural fidelity in RNA inverse design.

\section{Ablation Study}
\label{sec:ablation}

We conduct ablation studies on the advantage estimation strategy, reward design, and key hyperparameters used in both \texttt{RIDER} and \texttt{RIDE}.

\subsection{Advantage Estimation Strategy}
\label{sec:adv_ablation}

In \texttt{RIDER}, we replace raw rewards with advantage estimates and use the batch mean as a baseline for advantage calculation. Additionally, we introduce a moving average strategy to stabilize the baseline and reduce variance during training. In this section, we compare the full \texttt{RIDER} with two variants: (1) \textbf{\texttt{RIDER-RWD}}, which directly uses raw rewards without advantage estimation; and (2) \textbf{\texttt{RIDER-ADV}}, which uses batch-based advantage but without the moving average baseline.

Figures~\ref{fig:ADV_gdt}, \ref{fig:ADV_rmsd}, and \ref{fig:ADV_tm} show the performance of the three methods on the test set across three 3D self-consistency metrics. \texttt{RIDER} consistently achieves the best performance across all metrics. In contrast, \texttt{RIDER-RWD} performs poorly on many samples—for example, exhibiting numerous points with GDT\_TS scores below 0.1 or RMSD values exceeding 10\AA. This instability is likely due to the lack of advantage estimation, which causes high variance and sometimes leads to training collapse in later stages. 

Compared to \texttt{RIDER-ADV}, our full \texttt{RIDER} achieves around a 15\% improvement in GDT\_TS, demonstrating the effectiveness of the moving average strategy in stabilizing learning and enhancing structural quality.

Figures~\ref{fig:ADV_14_gdt}, \ref{fig:ADV_14_rmsd}, and \ref{fig:ADV_14_tm} further present the evaluation results of \texttt{RIDER}, \texttt{RIDER-RWD}, and \texttt{RIDER-ADV} on the 14 RNA structures of interest~\cite{das2010atomic}. On the GDT\_TS metric, \texttt{RIDER} achieves the best performance across 12 out of the 14 samples, and it consistently outperforms the other variants across all three evaluation metrics. For the remaining two samples, \texttt{RIDER} performs slightly worse than \texttt{RIDER-ADV} but still significantly outperforms \texttt{RIDER-RWD}.

\begin{figure*}[t!]
\center
    \includegraphics[width=0.7\linewidth]{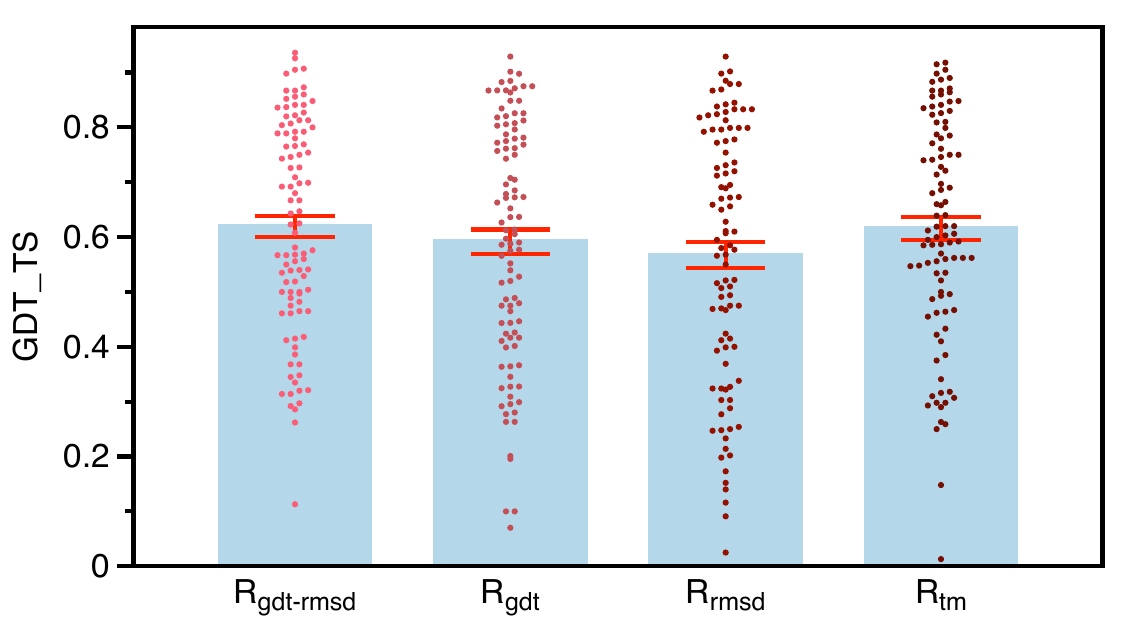}
    \caption{Comparison of four reward functions ($R^{\texttt{tm}}$, $R^{\texttt{rmsd}}$, $R^{\texttt{gdt}}$, $R^{\texttt{gdt\_rmsd}}$) on the GDT\_TS metric across the test set. Bar heights represent the mean, and red error bars indicate the SEM.}
    \label{fig:RWD_gdt}
\end{figure*}

\begin{figure*}[t!]
\center
    \includegraphics[width=0.7\linewidth]{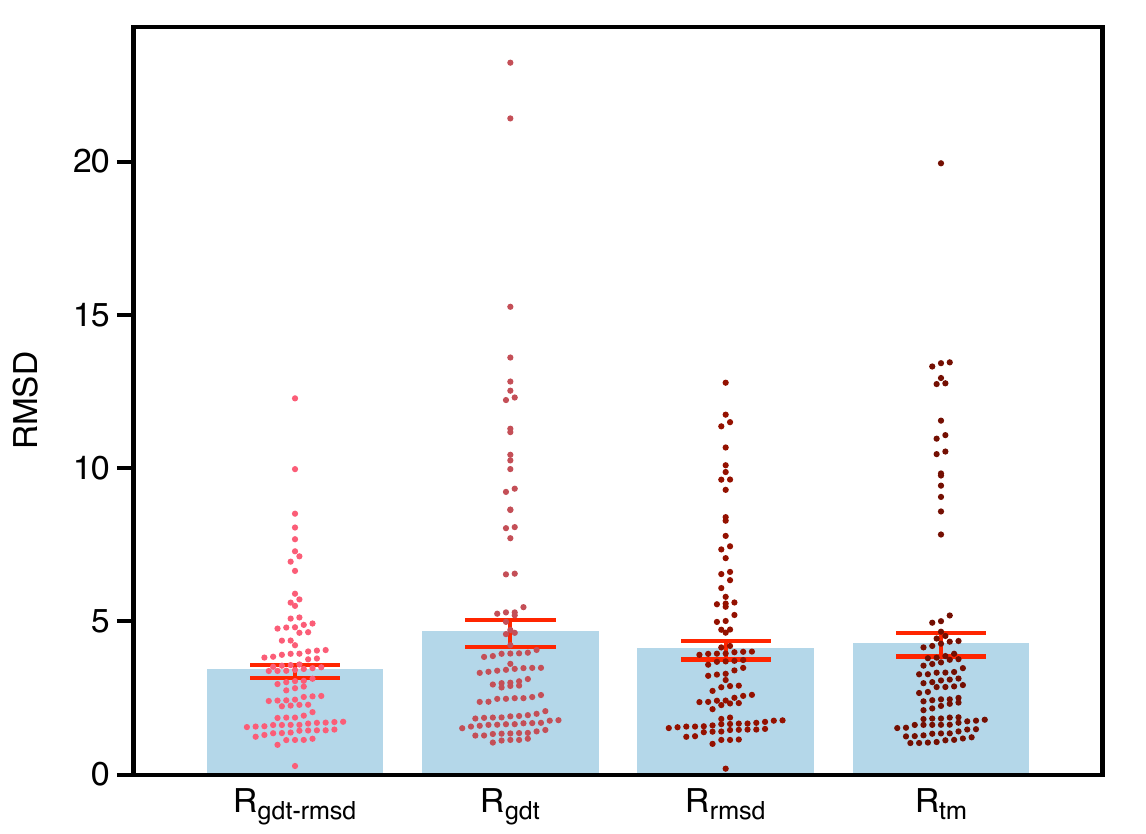}
    \caption{Comparison of four reward functions on the RMSD metric across the test set. Bar heights represent the mean, and red error bars indicate the SEM.}
    \label{fig:RWD_rmsd}
\end{figure*}

\begin{figure*}[t!]
\center
    \includegraphics[width=0.7\linewidth]{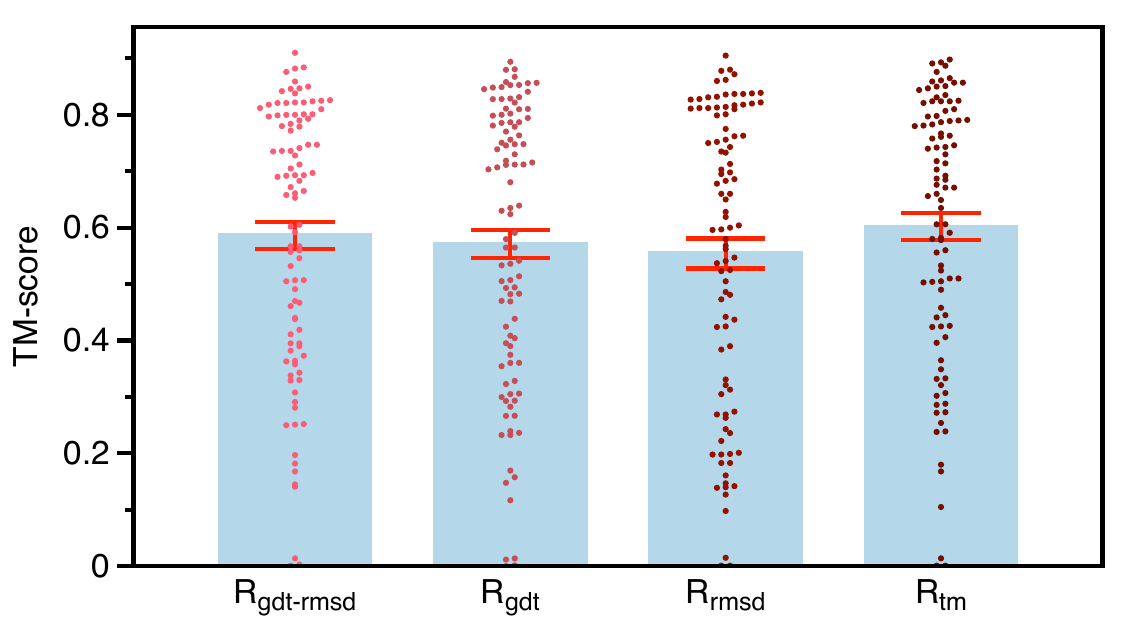}
    \caption{Comparison of four reward functions on the TM-score metric across the test set. Bar heights represent the mean, and red error bars indicate the SEM.}
    \label{fig:RWD_tm}
\end{figure*}

\subsection{Reward Functions}
\label{sec:rwd_ablation}

Figures~\ref{fig:RWD_gdt}, \ref{fig:RWD_rmsd}, and \ref{fig:RWD_tm} present the evaluation results of the four reward functions we designed, measured across all three structural similarity metrics on the test set.

Among them, the reward function based solely on TM-score, $R^{\texttt{tm}}$, performs well on both TM-score and GDT\_TS, but slightly worse on RMSD. The variants using only $R^{\texttt{rmsd}}$ or $R^{\texttt{gdt}}$ perform slightly worse than $R^{\texttt{tm}}$ overall.

We hypothesize that the inferior performance of $R^{\texttt{gdt}}$ may be due to the fact that GDT\_TS only considers well-modeled residues (i.e., those within 8\AA), and ignores errors in poorly modeled regions. This may cause the RL optimization to become trapped in local optima. On the other hand, while RMSD accounts for deviations across all residues, it does not apply length normalization like TM-score, making it potentially unsuitable for comparing RNAs of varying lengths. Moreover, in the course of exploration, some structures may yield very large RMSD values, leading to abrupt changes in reward, which can destabilize the training process.

In contrast, the composite reward $R^{\texttt{gdt\_rmsd}}$, which combines both GDT\_TS and RMSD, demonstrates strong performance across all metrics. We attribute this to the complementary nature of the two metrics—balancing local accuracy (GDT\_TS) with global deviation sensitivity (RMSD).

\begin{table}[t]
  \centering
  \caption{Ablation study of \texttt{RIDE} hyperparameters on native sequence recovery. Results are reported as the mean and standard deviation over five independent runs.}
  \label{tab:hyperparam-ablation}
  \resizebox{1\linewidth}{!}{
  \begin{tabular}{@{}llccc@{}}
    \toprule
    Optimizer & Activation function & Initial LR & \parbox{1in}{\centering GVP-GNN layers \\($L_E, L_D$)} & Native sequence recovery \\
    \midrule
    \multirow{8}{*}{\texttt{Adam}} & \multirow{4}{*}{\texttt{SiLU}} & \multirow{2}{*}{$3 \times 10^{-4}$} & 5 & $\mathbf{0.61 \pm 0.02}$ \\
                                     &                                  &                                   & 8 & $0.58 \pm 0.04$ \\
                                     \cmidrule(lr){3-5}
                                     &                                  & \multirow{2}{*}{$5 \times 10^{-4}$} & 5 & $0.60 \pm 0.02$ \\
                                     &                                  &                                   & 8 & $0.52 \pm 0.08$ \\
                                     \cmidrule(lr){2-5}
                                     & \multirow{4}{*}{\texttt{ReLU}} & \multirow{2}{*}{$3 \times 10^{-4}$} & 5 & $0.58 \pm 0.04$ \\
                                     &                                  &                                   & 8 & $0.59 \pm 0.03$ \\
                                     \cmidrule(lr){3-5}
                                     &                                  & \multirow{2}{*}{$5 \times 10^{-4}$} & 5 & $0.57 \pm 0.03$ \\
                                     &                                  &                                   & 8 & $0.51 \pm 0.04$ \\
    \midrule
    \multirow{8}{*}{\texttt{AdamW}} & \multirow{4}{*}{\texttt{SiLU}} & \multirow{2}{*}{$3 \times 10^{-4}$} & 5 & $0.59 \pm 0.03$ \\
                                      &                                  &                                   & 8 & $0.59 \pm 0.04$ \\
                                      \cmidrule(lr){3-5}
                                      &                                  & \multirow{2}{*}{$5 \times 10^{-4}$} & 5 & $0.47 \pm 0.10$ \\
                                      &                                  &                                   & 8 & $0.50 \pm 0.08$ \\
                                      \cmidrule(lr){2-5}
                                      & \multirow{4}{*}{\texttt{ReLU}} & \multirow{2}{*}{$3 \times 10^{-4}$} & 5 & $0.56 \pm 0.03$ \\
                                      &                                  &                                   & 8 & $0.53 \pm 0.09$ \\
                                      \cmidrule(lr){3-5}
                                      &                                  & \multirow{2}{*}{$5 \times 10^{-4}$} & 5 & $0.56 \pm 0.04$ \\
                                      &                                  &                                   & 8 & $0.50 \pm 0.10$ \\
    \bottomrule
  \end{tabular}
  }
\end{table}

\subsection{Hyperparameter Sensitivity of \texttt{RIDE}}
\label{sec:gnn_ablation}

We also perform a hyperparameter sensitivity analysis for the \texttt{RIDE} model. Table~\ref{tab:hyperparam-ablation} summarizes the native sequence recovery performance under different configurations of optimizer, activation function, initial learning rate, and the number of GVP-GNN layers ($L_E, L_D$). The results are averaged over five independent runs with different random seeds. 

The results show that the setting used in our main experiments—\texttt{Adam} optimizer, \texttt{SiLU} activation, initial learning rate of $3 \times 10^{-4}$, and 5 GVP-GNN layers—achieves the best sequence recovery ($0.61 \pm 0.02$). Deeper networks (8 layers) tend to underperform, possibly due to overfitting or optimization instability. We also observe that \texttt{SiLU} consistently outperforms \texttt{ReLU}, and that \texttt{AdamW} performs slightly worse than \texttt{Adam} under most configurations.

\subsection{Additional Results}
\label{sec:additional}

Figure~\ref{fig:rl_results} in Section~\ref{sec:experiments} presents the GDT\_TS comparison on 14 RNA structures of interest~\cite{das2010atomic} for gRNAde, \texttt{RIDE} (pre-trained), and \texttt{RIDER} (fine-tuned with $R^{\texttt{gdt\_rmsd}}$). Here, we provide the corresponding supplementary results on the other two metrics: RMSD and TM-score. 

\begin{figure*}[t!]
\center
    \includegraphics[width=1.0\linewidth]{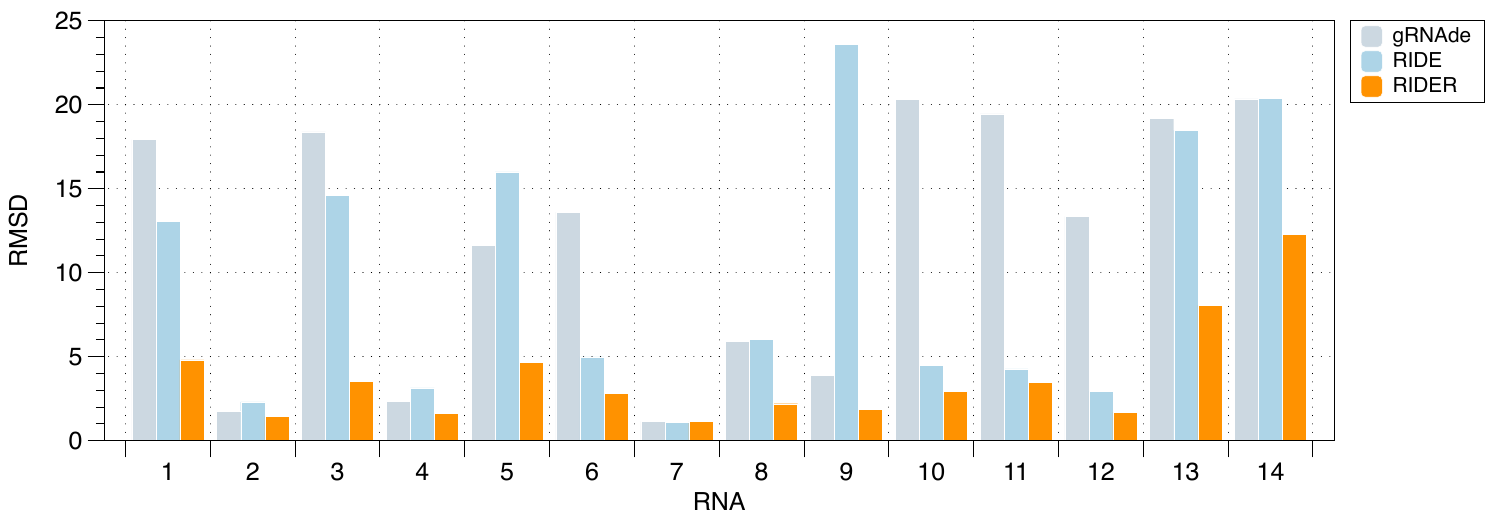}
    \caption{RMSD comparison of gRNAde, \texttt{RIDE}, and \texttt{RIDER} on 14 RNA structures of interest. Lower RMSD indicates better structural alignment.}
    \label{fig:comparsion_14_rmsd}
\end{figure*}

\begin{figure*}[t!]
\center
    \includegraphics[width=1.0\linewidth]{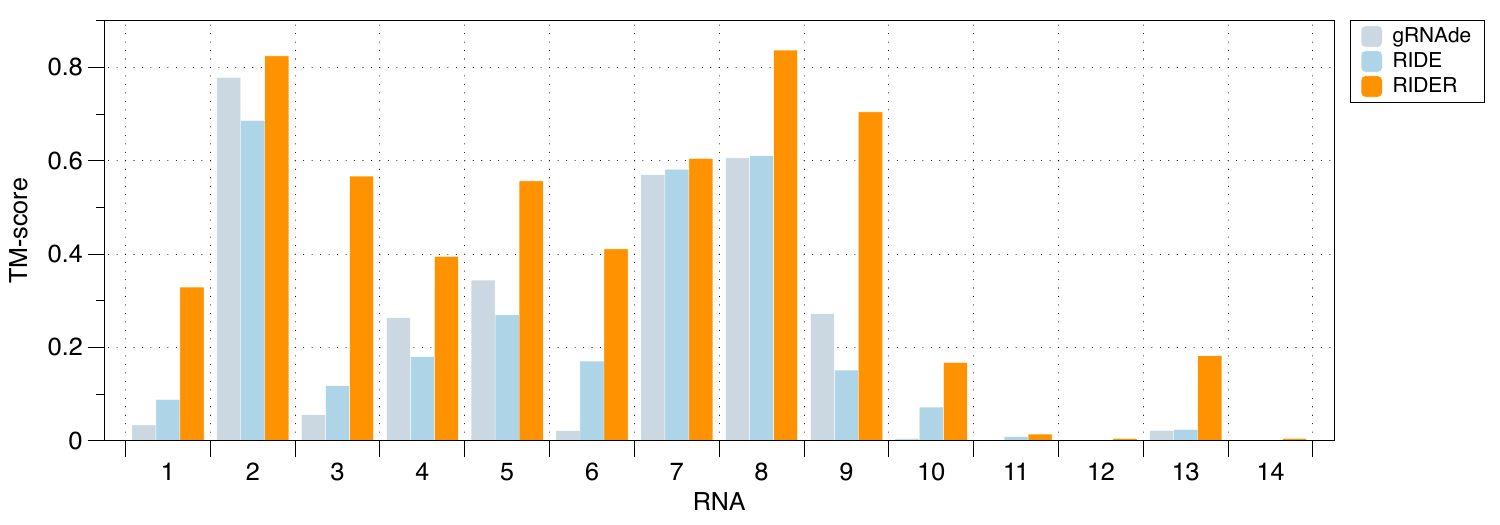}
    \caption{TM-score comparison of gRNAde, \texttt{RIDE}, and \texttt{RIDER} on 14 RNA structures of interest. Higher TM-score reflects better global fold similarity.}
    \label{fig:comparsion_14_tm}
\end{figure*}

Figure~\ref{fig:comparsion_14_rmsd} shows the RMSD comparison among the three methods, while Figure~\ref{fig:comparsion_14_tm} presents the TM-score comparison. These results further demonstrate that \texttt{RIDER} substantially outperforms both gRNAde and the pre-trained \texttt{RIDE} on the RMSD and TM-score metrics. In most cases, \texttt{RIDER} achieves improvements exceeding $100\%$, highlighting the effectiveness of our proposed method.

\section{Discussion}
\label{sec:discussion}

\subsection{Future work \& limitations}
\label{subsec:limitations}

The proposed \texttt{RIDER} framework is flexible and can be extended in several directions. Future work may explore the design of multi-objective reward functions that incorporate additional factors such as sequence diversity or functional properties. Another promising avenue is to integrate biophysical metrics, for example, energy-based terms, into the reward function. Such integration could guide the model towards sequences with improved thermodynamic stability, which is a critical requirement for practical applications.

Our current method focuses on \textit{in silico} RNA inverse design. However, RNA molecules are dynamic and can adopt multiple conformations \textit{in vivo}, further increasing the complexity of the inverse design task. As future work, we plan to experimentally validate our designed sequences and extend the framework to support multiple, potentially competing, optimization objectives.

\subsection{Broader Impacts}
\label{subsec:broader_impacts}

The proposed method may accelerate progress in synthetic biology, RNA-based drug development, and the study of gene regulatory mechanisms, thereby contributing to novel therapeutic strategies. By enabling the design of RNAs with precise 3D structures, it opens opportunities for engineering functional RNA circuits, aptamers, and ribozymes. However, applying AI models to biological sequence design introduces challenges related to interpretability and reliability. Enhancing the interpretability of such models—e.g., by identifying which structural features influence predictions—may help mitigate these concerns and promote responsible use in biological design.

\end{document}